\documentclass{article}
\PassOptionsToPackage{numbers,compress}{natbib}

\usepackage[preprint]{neurips_2026}

\usepackage[utf8]{inputenc} 
\usepackage[T1]{fontenc}    
\usepackage{hyperref}       
\usepackage{url}            
\usepackage{booktabs}       
\usepackage{amsfonts}       
\usepackage{nicefrac}       
\usepackage{microtype}      
\usepackage{xcolor}         

\usepackage{graphicx}
\usepackage{booktabs}   
\usepackage{multirow}   
\usepackage{enumitem}
\usepackage{amsmath}
\usepackage{algorithm}
\usepackage{algorithmic}
\usepackage{array}
\usepackage[table,dvipsnames]{xcolor}
\usepackage{colortbl}
\usepackage{wrapfig}
\usepackage{longtable}
\usepackage{array}

\usepackage{amsthm, amsmath, amssymb, booktabs}
\usepackage{tcolorbox}
\usepackage{subcaption} 
\usepackage{graphicx}  

\usepackage{xcolor}
\usepackage{tcolorbox}
\tcbuselibrary{theorems, skins, breakable}
\tcbuselibrary{theorems}

\usepackage{hyperref}

\definecolor{custom_blue}{HTML}{C5E0F0}
\definecolor{custom_blue_light2}{HTML}{F4FAFD}
\definecolor{custom_blue_light}{HTML}{E0EFF8}
\definecolor{custom_blue_dark}{HTML}{4A9DCB}
\definecolor{custom_blue_dark2}{HTML}{1F4E79}

\definecolor{darkblue}{rgb}{0, 0, 0.5}
\definecolor{c-green}{RGB}{52,181,145}
\definecolor{c-green-light}{RGB}{221,247,240}
\definecolor{c-green-dark}{RGB}{6,105,77}
\definecolor{c-pink-light}{RGB}{255,246,249}
\definecolor{c-pink}{RGB}{218,123,153}
\definecolor{c-pink-dark}{RGB}{179,62,122}

\newtcbtheorem[number within=section]{mytheorem}{Theorem}{
  colback=white,
  colframe=c-green-dark,
  enhanced,
  coltitle=black,
  colbacktitle=c-green-light,
  fonttitle=\small\bfseries,
  boxsep=2.7pt,
  top=1pt,
  bottom=1pt,
  left=4pt,
  right=4pt,
  separator sign={:},
}{thm}
\newtheorem{theorem}{Theorem}
\newtheorem{lemma}[theorem]{Lemma}

\definecolor{custom_green}{HTML}{ACE8D7}
\definecolor{custom_green_light2}{HTML}{EBF9F5}
\definecolor{custom_green_light}{HTML}{C8F6E9}
\definecolor{custom_green_dark}{HTML}{34B591}
\definecolor{custom_green_dark2}{HTML}{006246}
\definecolor{custom_pink}{HTML}{FFD2E0}
\definecolor{custom_pink_light2}{HTML}{FFF0F5}
\definecolor{custom_pink_light}{HTML}{FFE9F0}
\definecolor{custom_pink_dark}{HTML}{DE72AA}
\definecolor{custom_pink_dark2}{HTML}{962660}

\title{Rebellious Student: Reversing Teacher Signals for Reasoning Exploration with Self-Distilled RLVR}

\author{%
  Jeonghye Kim$^{1,2}$\thanks{Equal contribution. \textsuperscript{$\diamond$}Work done during an internship at Microsoft Research. \textsuperscript{$\dagger$} Corresponding author.} ~$^{\diamond}$, 
Jiwon Jeon$^{2*}$, Dongsheng Li$^1$, Yuqing Yang\textsuperscript{1$\dagger$} \\[4pt]
\textmd{$^1$Microsoft Research\, $^2$KAIST} \\
{\small \texttt{\{jeonghye.kim, jiwon.jeon\}@kaist.ac.kr}, \texttt{\{dongsli, yuqyang\}@microsoft.com}}
}

\begin{document}

\maketitle

\begin{abstract}
Self-distillation has emerged as a powerful framework for post-training LLMs, where a teacher conditioned on extra information guides a student without it, both from the same model. While this guidance is useful when the student has failed, on successful rollouts, the same mechanism instead overwrites the student's choices and suppresses it's own reasoning. Therefore, we propose reading the original self-distillation signal in reverse: when the student succeeds along a path the teacher would not have predicted, these tokens reflect its \emph{self-driven reasoning}. Building on this, we propose RLRT (RLVR with Reversed Teacher), which augments GRPO by reinforcing these tokens on correct rollouts. We interpret this as a new form of exploration in RLVR: not uniform diversity, but valuable exploration grounded in the student's own success. Across base, instruction-tuned, and thinking-tuned Qwen3 checkpoints, RLRT substantially outperforms self-distillation and exploration-based baselines, establishing information asymmetry as a new, principled design axis for RLVR.
\end{abstract}

\section{Introduction}
\label{sec:introduction}
\begin{figure}[h!]
    \centering
    \includegraphics[width=\linewidth]{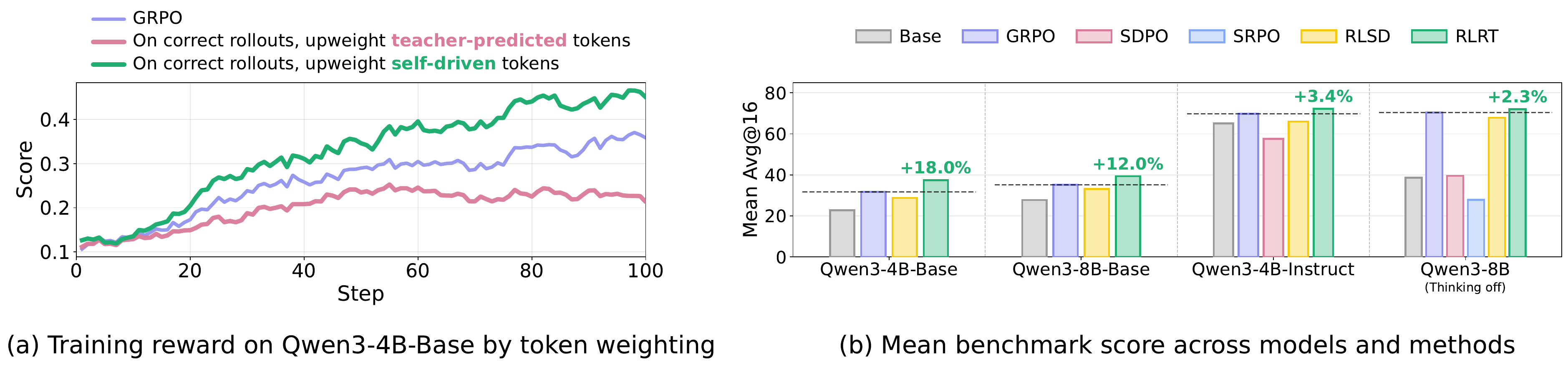}
    \caption{\small Reversing the teacher signal turns self-distillation into valuable exploration. (a) Training reward on Qwen3-4B-Base under \textcolor[HTML]{8C8EEE}{GRPO}, upweighting \textcolor{custom_pink_dark}{\textbf{teacher-predicted}} tokens on correct rollouts, and upweighting \textcolor{custom_green_dark}{\textbf{self-driven}} tokens (RLRT). (b) Mean avg@16 score over six math benchmarks (AIME24/25/26, HMMT26, AMC23, MATH500) across four Qwen3 backbones. RLRT consistently outperforms baselines significantly. Full results are in Tables \ref{tab:main_results} and \ref{tab:appendix_qwen3_4b_instruct}. We skipped detailed comparisons with SDPO/SRPO on the base models because they collapsed early during training (Appendix~\ref{appendix:sdpo_in_base}).}
    \label{fig:intro}
\end{figure}

Reinforcement learning with verifiable rewards (RLVR) has become the dominant paradigm for post-training LLMs on reasoning tasks~\citep{guo2025deepseekr1, shao2024deepseekmath}, yet it suffers from a credit-assignment bottleneck: the only learning signal is a sparse scalar reward at the end of each trajectory. Self-distillation has recently emerged as a powerful response~\citep{sdpo, sdft, opsd, rlsd}. Its core mechanism is an \emph{information asymmetry} between two views of the same model: a \emph{teacher} view conditioned on additional information (rich textual feedback, or a successful peer rollout) and a \emph{student} view without it. By distilling the teacher into the student, this asymmetry converts the sparse scalar reward into dense token-level supervision.

However, the value of distilling the teacher into the student depends on whether the rollout was already correct. On failed trajectories, conditioning the teacher on corrective information is useful: the teacher points the student toward solutions it could not previously reach on its own, and distillation transfers that corrective signal token by token. On already-successful trajectories, the same mechanism inverts its role. Even when the student already reached the correct answer, distilling toward the teacher overwrites the student's choices with the teacher's, a problem recently identified as \emph{optimization ambiguity} in self-distillation~\citep{srpo}. Rather than being corrected, the student is forced to imitate a path it had already solved its own way, undermining the independent reasoning that produced the success.

This observation motivates us to \emph{\textbf{reverse the direction}} of self-distillation on correct rollouts. Consider the tokens where the student's choice differs most sharply from what the teacher would have predicted. On a correct rollout, these are not arbitrary disagreements. They are the very points where the student exercised its own reasoning, choosing against the teacher and still arriving at the correct answer. Such tokens carry the student's \emph{self-driven reasoning}: choices that succeeded despite going against the teacher. Therefore, rather than suppressing them by aligning the student to the teacher, we propose to amplify these self-driven tokens during training. In this way, self-distillation becomes a tool for strengthening the student's reasoning ability, rather than reducing it to imitation.

This perspective also suggests a new angle for tackling the loss of reasoning diversity, a persistent failure mode of RLVR in which probability mass concentrates on trajectories the policy already prefers~\citep{yue2026does}. Existing methods address this through token-level entropy regulation~\citep{cui2025entropy, cliplowhigh, steer} or sequence-level diversity objectives~\citep{diver, dsdr, song2025outcome}, broadening exploration in the hope that wider sampling will surface correct paths. However, they treat diversity as a uniform target, leaving the RL signal to decide which alternative choices are worth keeping. We take a different stance. Rather than encouraging diversity for its own sake, we identify, within the rollouts the model has already produced, tokens that are simultaneously self-driven (departing from the conditioned teacher) and verified (occurring on correct trajectories), and upweight them during training. This yields what we term \emph{valuable exploration}: diversity grounded in successful reasoning rather than surface variation.

Building on this, we propose RLRT (RLVR with Reversed Teacher), which augments GRPO by reversing the direction of self-distillation on correct rollouts: instead of pulling the student to imitate the teacher, RLRT amplifies the self-driven tokens where the student reasoned differently from the teacher and still reached the correct answer. As shown in Figure~\ref{fig:intro}, across Qwen3-4B/8B-Base, Qwen3-4B-Instruct, and Qwen3-8B, RLRT exhibits faster training-score growth and outperforms self-distillation baselines by an average of \textbf{8.9\%} on six math reasoning benchmarks, including the challenging AIME and HMMT. We summarize our contributions as follows:

\begin{itemize}[itemsep=0pt, topsep=0pt, parsep=0pt, partopsep=0pt, leftmargin=*]
\item \textbf{A new analysis.} We reinterpret the teacher–student gap on correct rollouts: prior self-distillation reads it as an alignment target pulling the student to imitate the teacher, whereas we show that, read in reverse, it localizes the student's own \emph{self-driven reasoning}.
\item \textbf{A new algorithm.} Guided by this analysis, we propose \textbf{RLRT}, which augments GRPO by amplifying these self-driven tokens on correct rollouts, yielding consistent gains over strong RLVR baselines across base, instruction-tuned, and thinking-tuned models.
\item \textbf{A broader implication.} Beyond a specific algorithm, our findings establish \emph{information asymmetry} as a principled, intrinsic source of \emph{valuable exploration}, offering a new design axis for RLVR.
\end{itemize}

\section{Related Works} \label{sec:related}

\subsection{Self-Distillation in LLM Post-Training}
A growing line of work improves LLM reasoning through information asymmetry within a single model acting as both teacher and student, where the teacher is conditioned on privileged context. This context takes diverse forms: ground-truth reasoning traces~\citep{opsd}, runtime errors or judge evaluations as textual feedback~\citep{sdpo, empo2}, second-turn revisions conditioned on critiques~\citep{rltf}, expert demonstrations~\citep{sdft}, and prepended in-context knowledge or system prompts~\citep{opcd}. Across these variants, the design intent is alignment: the teacher–student gap is used to pull the student toward the teacher, whether by matching distributions~\citep{opsd, sdft, opcd}, distilling improved second-turn behavior into single-turn~\citep{rltf}, weighting tokens by the magnitude of teacher influence under verifiable rewards~\citep{rlsd}, or restricting alignment to failed rollouts only~\citep{srpo}. 

RLRT shares this asymmetric setup but inverts the alignment intent altogether: rather than pulling the student toward the teacher, we use the teacher–student gap in the opposite direction, treating tokens where the student diverged from the teacher on correct rollouts as evidence of self-driven reasoning, that is, choices made against the teacher's prediction that nonetheless reached the correct answer.

\subsection{Reasoning Exploration and Diversity}

RLVR is widely observed to suffer from reasoning boundary collapse, where the policy concentrates on a narrow set of high-reward strategies rather than expanding its reasoning capacity~\citep{yue2026does, reasoningboundaryparadox, rlvrdebate}. Existing remedies broaden output diversity at two scales: token-level entropy regulation~\citep{cui2025entropy, cliplowhigh, aepo, steer, reasoningwithexploration, positiveadvantage} and sequence- or outcome-level objectives over full reasoning traces~\citep{diver, dsdr, song2025outcome, passk, rrl}. Both treat diversity as a uniform target and rely on local stochasticity or heuristic proxies such as embedding similarity, n-gram overlap, or outcome counts, capturing surface variation rather than meaningful reasoning differences.

RLRT takes a different route. Rather than treating diversity as a uniform target, it identifies, within already-correct rollouts, the specific tokens at which the student departed from the teacher and yet still reached the correct answer, yielding valuable exploration: diversity grounded in the student's own successful reasoning rather than heuristic surrogates of variation.

\section{Preliminaries}
\label{sec:prelim}

\paragraph{Notation.}
Let $x$ be a prompt and $y = (y_1, \ldots, y_T)$ a response from policy $\pi_\theta$, with prefix $y_{<t} := (y_1, \ldots, y_{t-1})$ and suffix $y_{>t} := (y_{t+1}, \ldots, y_T)$. We write $h_t := (x, y_{<t})$ for the prefix history, $R \in \{0, 1\}$ for the verifiable reward, and $\mathcal{V}$ for the vocabulary.

\paragraph{Self-distillation in RLVR.}
In RLVR with self-distillation, a single model serves as both student and teacher: the student conditions only on $h_t$, while the teacher additionally conditions on a privileged context $c$ (e.g., the ground-truth solution or a successful rollout) hidden from the student~\citep{opsd, sdpo, rlsd}. We write
\begin{equation}
P_S^t(\cdot) \;:=\; \pi_\theta(\cdot \mid h_t),
\qquad
P_T^t(\cdot) \;:=\; \pi_\theta(\cdot \mid h_t, c),
\end{equation}
yielding a token-level log-probability ratio
$
\Delta_t \;:=\; \mathrm{sg}\!\left(\log P_T^t(y_t) - \log P_S^t(y_t)\right),
$
which measures how much the privileged context $c$ revises the model's belief about token $y_t$, with $\mathrm{sg}(\cdot)$ denoting stop-gradient.

Distribution-matching approaches such as on-policy self-distillation (OPSD)~\citep{opsd} use $\Delta_t$ to drive $P_S^t$ toward $P_T^t$ directly. RLSD~\citep{rlsd} observes that distribution matching is ill-posed when the student lacks access to $c$, since the target conditions on $c$ while the student does not. To avoid this, RLSD repurposes the ratio as a magnitude-only credit signal, yielding the RLSD update
\begin{equation}
w_t^{\mathrm{RLSD}} \;=\; \exp\!\bigl(\mathrm{sign}(A) \cdot \Delta_t\bigr) \;=\; \left(\frac{P_T^t(y_t)}{P_S^t(y_t)}\right)^{\mathrm{sign}(A)},
\label{eq:rlsd-weight}
\end{equation}
where $A$ is the group-relative advantage. The $\mathrm{sign}(A)$ exponent ensures direction-aware credit assignment: on correct rollouts, tokens with $P_T^t > P_S^t$ are amplified (the teacher \emph{favors} them); on incorrect rollouts, the same tokens are attenuated. Thus, the verifiable reward determines the sign of the update, while the teacher only modulates magnitude across tokens within a trajectory.

\section{Motivation}
\label{sec:motivation}

In RLVR, meaningful reasoning gains come not from rollouts that merely reach the correct answer, but from those that arrive there through novel paths, ones that diverge from the model's prior reasoning patterns. The teacher--student setup above provides a natural lens for identifying such moments. On correct rollouts, the tokens at which the student departs from the teacher are not merely mistakes to be suppressed, but signs of \emph{self-driven reasoning}. More formally, we identify self-driven reasoning with tokens at which the student deviates from the teacher's predictive distribution in ways influential to reaching the correct answer. Such tokens are what push the student toward stronger reasoning, and in this section we discuss how to detect and reinforce them.

\subsection{Information Asymmetry as an Exploration Signal}
\label{sec:motiv_ia}

To analyse self-driven reasoning, we define the \emph{token-level information asymmetry} $\hat{D}_t$ at a sampled token $y_t$ and the \emph{position-level information asymmetry} $\bar{D}_t$ as its expectation under the student:
\begin{equation}
\label{eq:ia-all}
\hat{D}_t(y_t) \;:=\; \log \frac{P_S^t(y_t)}{P_T^t(y_t)},
\qquad
\bar{D}_t \;:=\; \mathbb{E}_{v \sim P_S^t}[\hat{D}_t(v)] \;=\; \mathrm{KL}\bigl(P_S^t \,\big\|\, P_T^t\bigr).
\end{equation}

We claim that $\bar{D}_t$ flags \emph{which positions matter}, while the sign of $\hat{D}_t$ marks \emph{in which direction} the policy should update. 

\begin{figure}[h!]
    \centering
    \includegraphics[width=\linewidth]{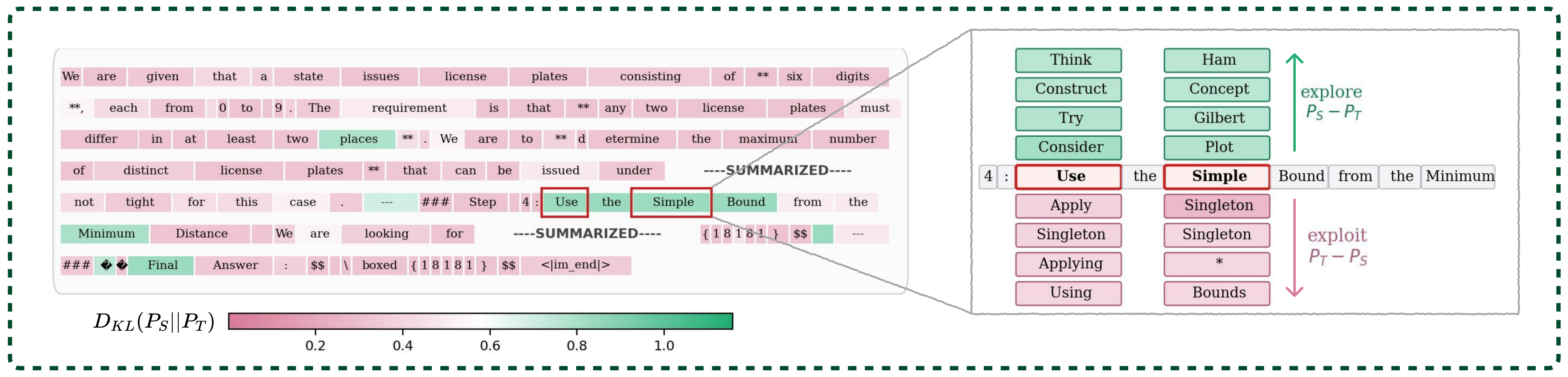}
    \caption{\textbf{Critical positions and explore/exploit directions.} Token shading shows the position-level asymmetry $\bar{D}_t = \mathrm{KL}(P_S^t \,\|\, P_T^t)$. At each critical position (right panels), candidate tokens are taken as the union of the teacher's and student's top-100 tokens; we display the top four with the largest $P_S^t - P_T^t$ (\textcolor{custom_green_dark}{green}, $\hat{D}_t > 0$) and the top four with the largest $P_T^t - P_S^t$ (\textcolor{custom_pink_dark}{pink}, $\hat{D}_t < 0$).}
    \label{fig:motivation}
\end{figure}

Figure~\ref{fig:motivation} illustrates $\bar{D}_t$ and $\hat{D}_t$ on a reasoning trajectory. Most tokens have small $\bar{D}_t$, but a few high-asymmetry tokens mark \emph{critical positions} where token choice strongly affects the outcome. At these positions, candidates the teacher would have predicted ($\hat{D}_t < 0$, e.g., \emph{use}, \emph{conclude}) define the \emph{exploit} direction, while candidates the student chose against the teacher's prediction ($\hat{D}_t > 0$, e.g., \emph{try}, \emph{consider}) define the \emph{explore} direction. Additional rollouts exhibiting the same pattern are provided in Appendix~\ref{app:more-examples}. In the following subsections, we examine $\bar{D}_t$ and $\hat{D}_t$ in more detail.

\subsection{$\bar{D}_t$ Identifies Which Positions Matter}
\label{sec:motiv_barDt}
\paragraph{Claim.}
The position-level information asymmetry $\bar{D}_t$ is large precisely at positions where the choice of token meaningfully affects the probability of a correct outcome.

\paragraph{Theoretical Justification.}
We justify the claim through a Bayesian view of the teacher. We model the teacher as $\pi_\theta$ conditioned on the event $R=1$ (success), so that the student and teacher distributions become
\begin{equation}
\label{eq:def_pspt}
P_S^t(\cdot) := \pi_\theta(\cdot \mid h_t),
\qquad
P_T^t(\cdot) := \pi_\theta(\cdot \mid h_t,\, R=1).
\end{equation}
For each token $v \in \mathcal{V}$, let
\[
f(v) \;:=\; \Pr_{Y \sim P_S}[R = 1 \mid h_t,\, y_t = v],
\qquad
\bar{f}_S^t \;:=\; \mathbb{E}_{v \sim P_S^t}[f(v)],
\]
denote the per-token correctness probability and its student-mean. Bayes' rule then yields a single identity that underlies the analysis below.

\begin{lemma}[Bayesian teacher]
\label{lem:bayes-teacher}
At each step $t$,
$P_T^t(v) = \tfrac{P_S^t(v)f(v)}{\bar{f}_S^t} \iff \hat{D}_t(v) = \log\bar{f}_S^t - \log f(v).$
\end{lemma}

\noindent The proof is deferred to Appendix~\ref{app:proof-bayes}. The teacher is the student tilted toward tokens with higher $f(v)$; equivalently, $\hat{D}_t(v)$ measures how far $f(v)$ falls below $\bar{f}_S^t$.

In RLVR, any policy update at position $t$ acts only on tokens the student actually samples, so the relevant signal is how much $f$ varies among such tokens. We call this the \emph{influence} of position $t$:
\begin{equation}
\mathrm{Inf}_S(t) \;:=\; \mathbb{E}_{v \sim P_S^t}\!\bigl[\,\bigl|f(v) - \bar{f}_S^t\bigr|\,\bigr].
\label{eq:inf-def}
\end{equation}
A position is \emph{critical} when $\mathrm{Inf}_S(t)$ is large and \emph{inert} when near zero. While $\hat{D}_t(y_t)$ acts pointwise, its student-expectation $\bar{D}_t = \mathrm{KL}(P_S^t \,\|\, P_T^t)$ from \eqref{eq:ia-all} captures the per-position effect of reweighting. The two scales are tied by a Pinsker-type bound.

\begin{theorem}[$\bar{D}_t$ controls $\mathrm{Inf}_S(t)$]
\label{thm:master}
At every step $t$, $\mathrm{Inf}_S(t)^2 \;\le\; 2\,\bar{D}_t$.
\end{theorem}

\noindent By contrapositive, $\bar{D}_t \approx 0$ implies $\mathrm{Inf}_S(t) \approx 0$: small asymmetry guarantees an inert position. The proof bounds $\mathrm{Inf}_S(t)$ by total variation distance using Lemma~\ref{lem:bayes-teacher}, then applies Pinsker's inequality (Appendix~\ref{app:proof-master}).

\subsection{Sign of $\hat{D}_t$ Identifies Which Direction to Push}
\label{sec:motiv_signDt}

At a critical position, the sign of $\hat{D}_t(y_t)$ determines which way to push. Two regimes follow directly from the definition $\hat{D}_t(v) := \log P_S^t(v) - \log P_T^t(v)$:
\begin{itemize}[itemsep=0pt, topsep=0pt, parsep=0pt, partopsep=0pt, leftmargin=*]
    \item $\hat{D}_t(v) < 0$: the token $v$ is more likely under the teacher ($P_T^t > P_S^t$), a choice the teacher would have predicted. Reinforcing such tokens follows the teacher's path, the \textbf{\textcolor{custom_pink_dark}{\emph{exploit}}} direction.
    \item $\hat{D}_t(v) > 0$: conversely, $v$ is a choice against the teacher's prediction ($P_S^t > P_T^t$). Reinforcing such tokens moves the student onto a self-driven path consistent with success, the \textbf{\textcolor{custom_green_dark}{\emph{explore}}} direction.
\end{itemize}

\noindent While the analysis above defines the teacher through the abstract event $R=1$, this event cannot be conditioned on directly. In practice, we realize the teacher by feeding a known correct solution $c$ as the conditioning context, so that $P_T^t(\cdot) = \pi_\theta(\cdot \mid h_t,\, c)$ serves as one instantiation of $\pi_\theta(\cdot \mid h_t,\, R=1)$.

\begin{figure}[h!]
    \centering
    \includegraphics[width=\linewidth]{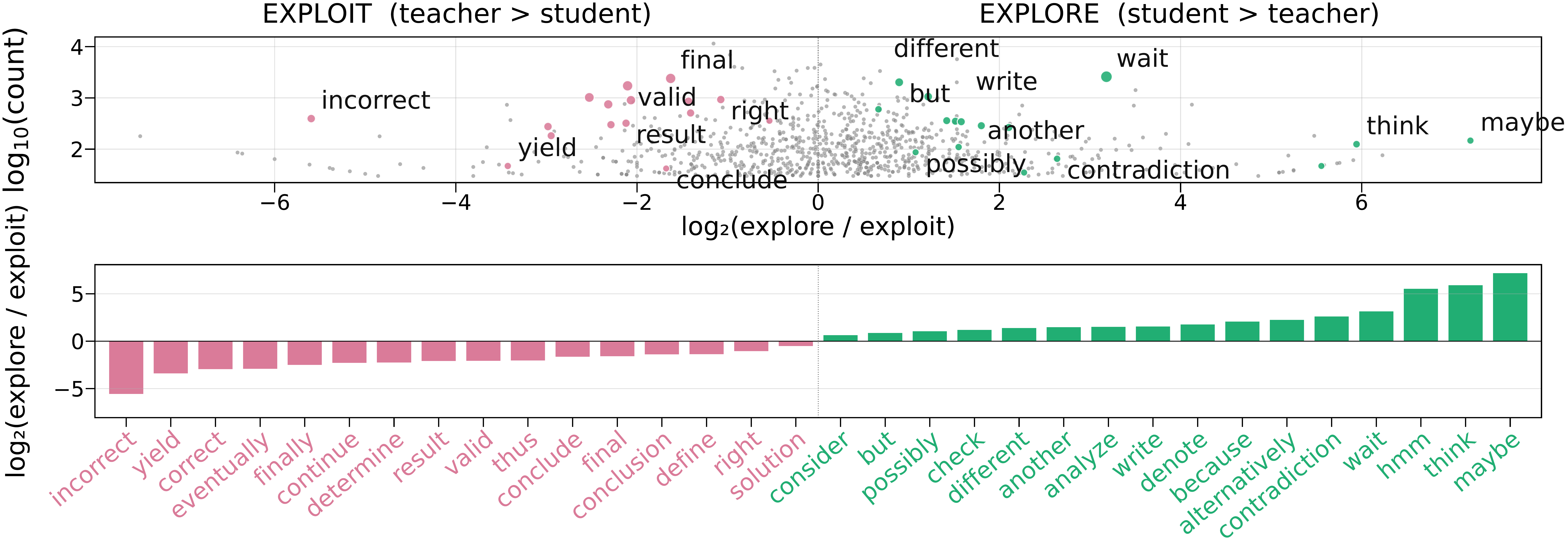}
    \caption{\textbf{Reasoning markers in the explore/exploit population.}
    (a)~Volcano scatter of linguistic tokens: $x$-axis is the polarization $\log_{2}(n_{\text{explore}}/n_{\text{exploit}})$, $y$-axis is the total count $\log_{10}(n_{\text{explore}}+n_{\text{exploit}})$. Highlighted points (\textcolor{custom_green_dark}{green} $=$ explore, \textcolor{custom_pink_dark}{pink} $=$ exploit) are categorized discourse markers; grey points are uncategorized tokens. (b)~Per-marker polarization for these markers, sorted from most exploit-leaning (left) to most explore-leaning (right).}
    \label{fig:markers}
\end{figure}

To verify that the sign of $\hat{D}_t$ captures the explore/exploit direction, we ask which tokens the student systematically chooses against the teacher's prediction versus which tokens align with it across rollouts from Qwen3-8B on DAPO-Math-17k \citep{dapo}. We score each token's polarization between the two sides with the smoothed log-odds $z$-score of \citet{monroe2008fightin}. Figure~\ref{fig:markers} shows that explore-leaning tokens open new reasoning paths (\emph{wait}, \emph{another}, \emph{consider}), while exploit-leaning tokens close them with verdicts and conclusions (\emph{conclude}, \emph{correct}, \emph{final}). Full details of the marker selection and the per-category list are provided in Appendix~\ref{appendix:marker-stats}.

\section{RLRT: RLVR with Reversed Teacher}
\label{sec:method}

\begin{figure}[h!]
    \centering
    \begin{subfigure}[b]{0.36\linewidth}
        \centering
        \raisebox{0.25cm}{\includegraphics[width=\linewidth]{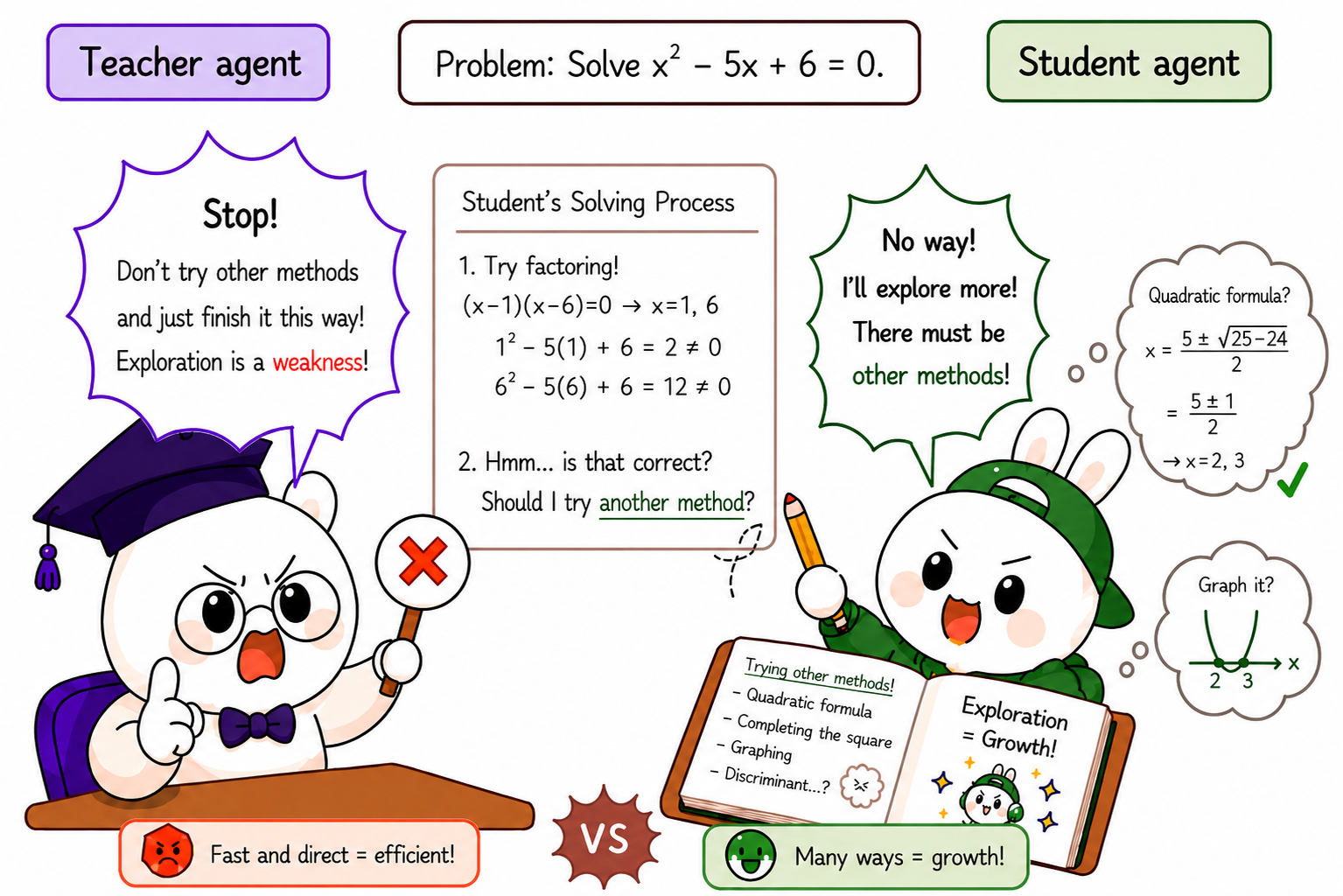}}
        \caption{Conceptual illustration.}
        \label{fig:rlrt_concept}
    \end{subfigure}
    \hfill
    \begin{subfigure}[b]{0.62\linewidth}
        \centering
        \includegraphics[width=\linewidth]{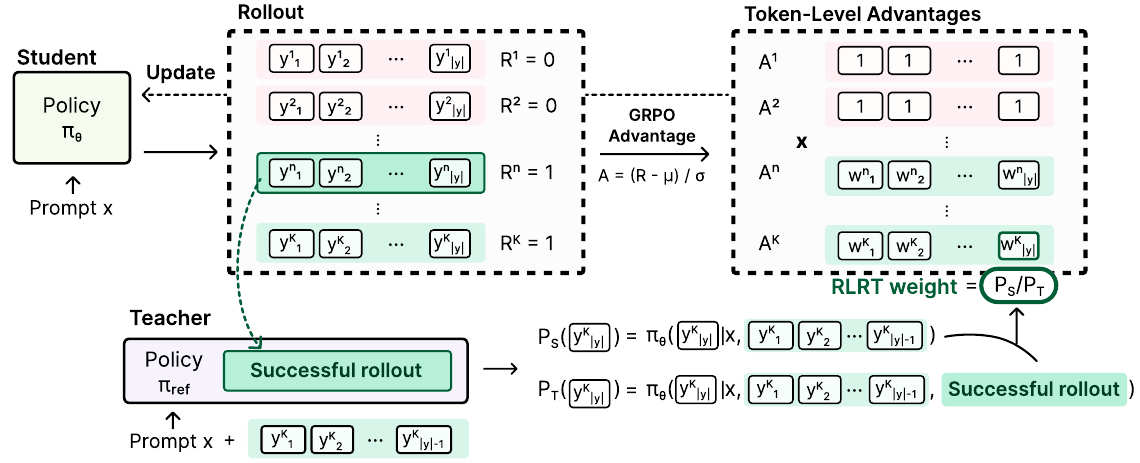}
        \caption{Training pipeline.}
        \label{fig:rlrt_pipeline}
    \end{subfigure}
    \caption{\textbf{Overview of RLRT.} 
    \textbf{(a)} Conceptual illustration of the reversed-teacher signal.
    \textbf{(b)} Given a prompt $x$, the student policy $\pi_\theta$ generates $K$ rollouts that receive verifiable rewards $r \in \{0,1\}$ and group-standardized advantages $A^{(k)}$. A reversed teacher provides token-level signals $\hat{D}_t$ that, on correct rollouts, up-weight tokens with $\hat{D}_t > 0$.}
    \label{fig:rlrt_overview}
\end{figure}

We now present \textbf{RLRT} (RLVR with Reversed Teacher), an instance of the framework in Section~\ref{sec:motivation} that uses an informed teacher and amplifies, on correct rollouts, tokens with $\hat{D}_t > 0$. RLRT modifies only the token-level credit assignment of standard GRPO~\citep{shao2024deepseekmath}, leaving the rollout, reward, and trust-region machinery unchanged. Figure~\ref{fig:rlrt_overview} provides a conceptual illustration and the training pipeline of RLRT.

\paragraph{Reverse Weight as Token-Level Information Asymmetry Credit.}
For a prompt $x$, the student policy $\pi_\theta$ samples a group of $K$ rollouts $\{y^{(k)}\}_{k=1}^K$, each receiving a verifiable reward $r(y^{(k)}) \in \{0,1\}$ and a group-standardized advantage $A^{(k)}$. RLRT defines a per-token reweighting based on $\hat{D}_t$:
\begin{equation}
w_t^{\mathrm{RLRT}} \;=\; \exp\!\bigl(\mathrm{sign}(A) \cdot \hat{D}_t\bigr) \;=\; \left(\frac{P_S^t(y_t)}{P_T^t(y_t)}\right)^{\mathrm{sign}(A)}.
\label{eq:rlrt-weight}
\end{equation}
On positive-advantage tokens, $w_t^{\mathrm{RLRT}} > 1$ exactly when $\hat{D}_t > 0$, i.e., for tokens the student chose against the teacher's prediction, and the reweighting amplifies these self-driven choices rather than aligning the student to the teacher. The flipping of the teacher/student ratio relative to the RLSD update~\citep{rlsd} (Eq.~\ref{eq:rlsd-weight}) reflects a difference in intent: RLSD treats teacher--student disagreement as a correction to be applied, whereas RLRT treats it as a signal of valuable exploration and amplifies it.

\paragraph{Reward-Gated Update.}
Following the framework's requirement that token-level information asymmetry be combined with outcome conditioning to target self-driven tokens on correct trajectories, the reverse weight is applied only to correct rollouts:
\begin{equation}
A_t^{\mathrm{RLRT},(k)} \;=\;
\begin{cases}
A^{(k)} \cdot \Bigl[\,(1-\lambda) + \lambda \cdot \mathrm{clip}\bigl(w_t^{\mathrm{RLRT}},\,1-\varepsilon_w,\,1+\varepsilon_w\bigr)\Bigr] & \text{if } r(y^{(k)}) = 1, \\[4pt]
A^{(k)} & \text{if } r(y^{(k)}) = 0,
\end{cases}
\label{eq:rlrt-advantage}
\end{equation}
where $\lambda \in [0,1]$ controls the strength of the reversed signal ($\lambda = 0$ recovers vanilla GRPO, $\lambda = 1$ yields full reverse weighting), and the clip $\varepsilon_w$ bounds the per-token advantage perturbation by $\lambda \cdot \varepsilon_w$.

\section{Experiments} \label{sec:experiments}

We design our experiments to verify that RLRT effectively leverages the information asymmetry signal to induce \emph{valuable exploration} 
during RLVR training. Concretely, we ask:
\begin{itemize}[itemsep=0pt, topsep=0pt, parsep=0pt, partopsep=0pt, leftmargin=*]
    \item \textbf{(Q1)} How does RLRT, which pushes the student \emph{away from} the teacher on correct rollouts, perform compared to self-distillation methods that pull the student \emph{toward} the teacher?
    \item \textbf{(Q2)} Does $\bar{D}_t$ causally identify critical positions, and does RLRT amplify their effect?
    \item \textbf{(Q3)} Beyond sharpening the base's confident predictions, does RLRT introduce meaningful change?
    \item \textbf{(Q4)} Does RLRT induce more effective exploration than prior exploration-based methods?
\end{itemize}
 
\subsection{Benchmark Results} \label{sec:benchmark_results}

\paragraph{Experimental Setup.} To answer (Q1), we use DAPO-Math-17k~\citep{dapo} as the training corpus. Since post-training dynamics depend strongly on the pretrained checkpoint's inductive biases~\citep{zhao2025echo, zhang2025interplay}, we evaluate on three qualitatively distinct model types: a \textbf{base} model (Qwen3-4B/8B-Base), an \textbf{instruction-tuned} model (Qwen3-4B-Instruct), and a \textbf{thinking-tuned} model (Qwen3-8B). 

We compare RLRT against GRPO and three self-distillation baselines, SDPO~\citep{sdpo}, SRPO~\citep{srpo}, and RLSD~\citep{rlsd}. We adopt SDPO rather than the closely related OPSD~\citep{opsd}, since OPSD relies on ground-truth solutions from an external dataset and on a hybrid setup in which the student runs with thinking disabled and the teacher with thinking enabled. SDPO instead operates entirely on the model's own rollouts, consistent with our self-distillation setup.
Details of each algorithm are provided in Appendix~\ref{appendix:baseline_details}. In addition, SDPO collapsed early on Qwen3-4B/8B-Base (Appendix~\ref{appendix:sdpo_in_base}), so we omit a detailed comparison for base models. We use a training batch size of 256, a PPO mini-batch size of 128, and a maximum response length of 20{,}480 tokens, with asymmetric clipping $\varepsilon_{\text{high}}{=}0.28$ and $\varepsilon_{\text{low}}{=}0.2$ following~\citet{dapo}. Further hyperparameters are listed in Appendix~\ref{app:hyperparams}.

\begin{figure}[t!]
    \centering
    \includegraphics[width=\linewidth]{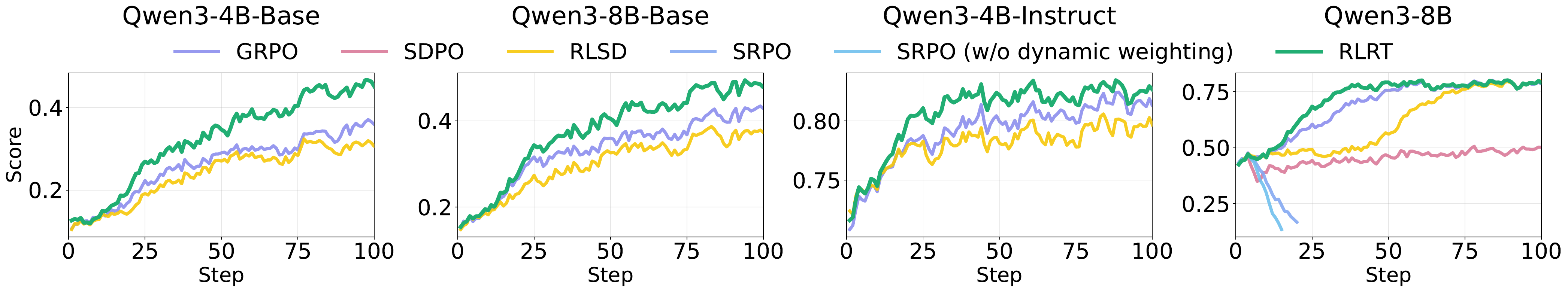}
    \caption{\small Training score curves across four backbones (Qwen3-4B-Base, Qwen3-8B-Base, 
Qwen3-4B-Instruct, Qwen3-8B). RLRT achieves faster exploration and 
higher training scores in all settings.}
    \label{fig:training_curves}
\end{figure}

\newcolumntype{C}[1]{>{\centering\arraybackslash}p{#1}}

\begin{table}[t!]
    \centering
    \small
    \setlength{\tabcolsep}{3pt}
    \renewcommand{\arraystretch}{1.15}
    \caption{Performance comparison across mathematical reasoning benchmarks. We report avg@16 and pass@16 for each benchmark. $\Delta$ denotes the gain of RLRT over the best of the other methods. Due to space constraints, results for Qwen3-4B-Instruct are in Table \ref{tab:appendix_qwen3_4b_instruct} (Appendix \ref{appendix:qwen3_4b_instruct}).}
    \resizebox{\textwidth}{!}{%
    \begin{tabular}{p{2.55cm} C{0.9cm}C{0.9cm} C{0.9cm}C{0.9cm} C{0.9cm}C{0.9cm} C{0.9cm}C{0.9cm} C{0.9cm}C{0.9cm} C{0.9cm}C{0.9cm}}
        \toprule
        \multirow{2}{*}{\textbf{Method}} 
            & \multicolumn{2}{c}{\textbf{AIME24}} 
            & \multicolumn{2}{c}{\textbf{AIME25}} 
            & \multicolumn{2}{c}{\textbf{AIME26}} 
            & \multicolumn{2}{c}{\textbf{HMMT26}} 
            & \multicolumn{2}{c}{\textbf{AMC23}} 
            & \multicolumn{2}{c}{\textbf{MATH500}} \\
        \cmidrule(lr){2-3} \cmidrule(lr){4-5} \cmidrule(lr){6-7} \cmidrule(lr){8-9} \cmidrule(lr){10-11} \cmidrule(lr){12-13}
             & {\scriptsize Avg@16} & {\scriptsize Pass@16} 
            & {\scriptsize Avg@16} & {\scriptsize Pass@16} 
            & {\scriptsize Avg@16} & {\scriptsize Pass@16} 
            & {\scriptsize Avg@16} & {\scriptsize Pass@16} 
            & {\scriptsize Avg@16} & {\scriptsize Pass@16} 
            & {\scriptsize Avg@16} & {\scriptsize Pass@16} \\
        \midrule
       \textit{Qwen3-4B-Base}             & 9.6 & 33.3 & 6.9 & 30.0 & 6.5 & 16.7 & 3.6 & 24.2 & 43.3 & 90.0 & 66.8 & 92.2 \\
        \quad GRPO                         & 15.0 & 40.0 & 14.4 & 33.3 & 12.3 & 36.7 & 10.0 & 27.3 & 58.3 & 87.5 & 80.2 & \textbf{94.2} \\
        \quad RLSD                         & 13.3 & 40.0 & 11.2 & 33.3 & 9.0 & 26.7 & 6.2 & 27.3 & 55.2 & 82.5 & 77.9 & 91.2 \\
        \quad \textbf{RLRT (Ours)}         & \textbf{22.5} & \textbf{50.0} & \textbf{18.5} & \textbf{36.7} & \textbf{19.8} & \textbf{40.0} & \textbf{15.9} & \textbf{33.3} & \textbf{63.9} & \textbf{95.0} & \textbf{83.8} & \textbf{94.2} \\
        \quad $\Delta$ vs. best            & \cellcolor{gray!10}\textcolor{teal}{+7.5} & \cellcolor{gray!10}\textcolor{teal}{+10.0} & \cellcolor{gray!10}\textcolor{teal}{+4.1} & \cellcolor{gray!10}\textcolor{teal}{+3.4} & \cellcolor{gray!10}\textcolor{teal}{+7.5} & \cellcolor{gray!10}\textcolor{teal}{+3.3} & 
        \cellcolor{gray!10}\textcolor{teal}{+5.9} & 
        \cellcolor{gray!10}\textcolor{teal}{+6.0} & 
        \cellcolor{gray!10}\textcolor{teal}{+5.6} & \cellcolor{gray!10}\textcolor{teal}{+5.0} & \cellcolor{gray!10}\textcolor{teal}{+3.6} & \cellcolor{gray!10}{0.0} \\
        \midrule
        \textit{Qwen3-8B-Base}             & 10.4 & 33.3 & 10.2 & 30.3 & 9.8 & 30.0 & 5.3 & 30.3 & 56.3 & 85.0 & 74.4 & 93.0 \\
        \quad GRPO                         & 19.8 & 40.0 & 17.5 & 36.7 & 16.5 & 36.7 & 11.0 & \textbf{33.3} & 62.5 & 90.0 & 83.6 & 95.4 \\
        \quad RLSD                         & 17.3 & 40.0 & 15.0 & 33.3 & 13.5 & 36.7 & 8.1 & 27.3 & 64.2 & 87.5 & 80.9 & 92.4 \\
        \quad \textbf{RLRT (Ours)}         & \textbf{27.9} & \textbf{63.3} & \textbf{18.8} & \textbf{50.0} & \textbf{21.9} & \textbf{53.3} & \textbf{15.9} & \textbf{33.3} & \textbf{67.3} & \textbf{97.5} & \textbf{84.4} & \textbf{95.6} \\
        \quad $\Delta$ vs. best            & \cellcolor{gray!10}\textcolor{teal}{+8.1} & \cellcolor{gray!10}\textcolor{teal}{+23.3} & \cellcolor{gray!10}\textcolor{teal}{+1.3} & \cellcolor{gray!10}\textcolor{teal}{+13.3} & \cellcolor{gray!10}\textcolor{teal}{+5.4} & \cellcolor{gray!10}\textcolor{teal}{+16.6} & 
        \cellcolor{gray!10}\textcolor{teal}{+4.9} & 
        \cellcolor{gray!10}{0.0} & 
        \cellcolor{gray!10}\textcolor{teal}{+3.1} & \cellcolor{gray!10}\textcolor{teal}{+7.5} & \cellcolor{gray!10}\textcolor{teal}{+0.8} & \cellcolor{gray!10}\textcolor{teal}{+0.2} \\
        \midrule
        \textit{Qwen3-8B \tiny{(Thinking off)}}   & 25.2 & 63.3 & 20.0 & 43.3 & 15.4 & 50.0 & 20.3 & 33.3 & 67.0 & 95.0 & 83.7 & 95.8 \\
        \quad GRPO                          & 70.2 & 86.7 & 59.4 & 83.3 & 62.9 & \textbf{86.7} & 41.7 & 66.7 & 93.6 & \textbf{100.0} & 94.8 & \textbf{98.2} \\
        \quad SDPO                         & 26.9 & 63.3 & 22.3 & 40.0 & 14.4 & 36.7 & 18.6 & 30.3 & 72.8 & 95.0 & 82.2 & 94.4 \\
        \quad SRPO                         & 15.4 & 26.7 & 9.8 & 26.7 & 8.3 & 26.7 & 9.7 & 21.2 & 49.2 & 77.5 & 75.0 & 90.2 \\
        \quad RLSD                         & 65.4 & 83.3 & 57.9 & 83.3 & 57.7 & 83.3 & 39.2 & 51.5 & 93.6 & \textbf{100.0} & 93.1 & \textbf{98.2} \\
        \quad \textbf{RLRT (Ours)}         & \textbf{70.6} & \textbf{93.3} & \textbf{62.9} & \textbf{86.7} & \textbf{65.0} & \textbf{86.7} & \textbf{43.2} & \textbf{69.7} & \textbf{95.5} & \textbf{100.0} & \textbf{94.9} & \textbf{98.2} \\
        \quad $\Delta$ vs. best            & \cellcolor{gray!10}\textcolor{teal}{+0.4} & \cellcolor{gray!10}\textcolor{teal}{+6.6} & \cellcolor{gray!10}\textcolor{teal}{+3.5} & \cellcolor{gray!10}\textcolor{teal}{+3.4} & \cellcolor{gray!10}\textcolor{teal}{+2.1} & \cellcolor{gray!10}{0.0} & \cellcolor{gray!10}\textcolor{teal}{+1.5} & \cellcolor{gray!10}\textcolor{teal}{+3.0} & \cellcolor{gray!10}\textcolor{teal}{+1.9} & \cellcolor{gray!10}{0.0} &
        \cellcolor{gray!10}\textcolor{teal}{+0.1} &
        \cellcolor{gray!10}{0.0} \\
        \bottomrule
    \end{tabular}%
    }
    \label{tab:main_results}
\end{table}

\paragraph{Performance Comparison.} Figure~\ref{fig:training_curves} shows the training curves for each algorithm, and Table~\ref{tab:main_results} presents the evaluation results of the trained models on six math benchmarks using avg@16 and pass@16. As shown in Figure \ref{fig:training_curves} and Table \ref{tab:main_results}, across all four backbones, RLRT substantially outperforms both GRPO and the self-distillation baselines, exhibiting faster training-score growth and yielding significant average benchmark gains of \textbf{18.0\%} (Qwen3-4B-Base), \textbf{12.0\%} (Qwen3-8B-Base), \textbf{3.4\%} (Qwen3-4B-Instruct), and \textbf{2.2\%} (Qwen3-8B) over the baselines. Notably, SRPO, which routes correct rollouts to GRPO and incorrect rollouts to self-distillation, performs even worse than full self-distillation on math. We conjecture that self-distillation and GRPO promote different reasoning styles (e.g., exploration and exploitation as discussed in Section \ref{sec:motiv_signDt}), leading to conflicting gradients.
The gain is largest on Qwen3-4B-Base and smallest on Qwen3-8B, suggesting that RLRT's exploration signal is most effective when the policy has not yet been concentrated by instruction tuning. 

\subsection{Causal Intervention via Reflection Injection}
\label{subsec:val-critical}

We answer (Q2) by injecting the reflection prompt 
\emph{``Wait, let me reconsider.''} at a chosen token in a rollout 
and letting the model continue: if high-$\bar{D}_t$ tokens are 
truly critical branch points, this should flip outcomes there 
more often than elsewhere. We run this on $100$ DAPO-Math-17k 
problems ($8$ rollouts each) across \texttt{Qwen3-8B} checkpoints 
from step~$0$ (base) to step~$100$ under both RLRT and GRPO, 
injecting at three positions: $\arg\max_t \bar{D}_t$ 
(\texttt{max\_kl}), a uniform-random token 
(\texttt{random}), and $\arg\min_t \bar{D}_t$ (\texttt{min\_kl}). On the hard ($n_\mathrm{correct} \in \{0,1,2\}$) 
and easy ($\{5,6,7\}$) subsets, we report \emph{flip$\to$R} 
(wrong$\to$right) and \emph{flip$\to$W} (right$\to$wrong) rates, 
respectively.

\begin{wrapfigure}{r}{0.415\textwidth}
    \centering
    \vspace{-0.3cm}
    \includegraphics[width=\linewidth]{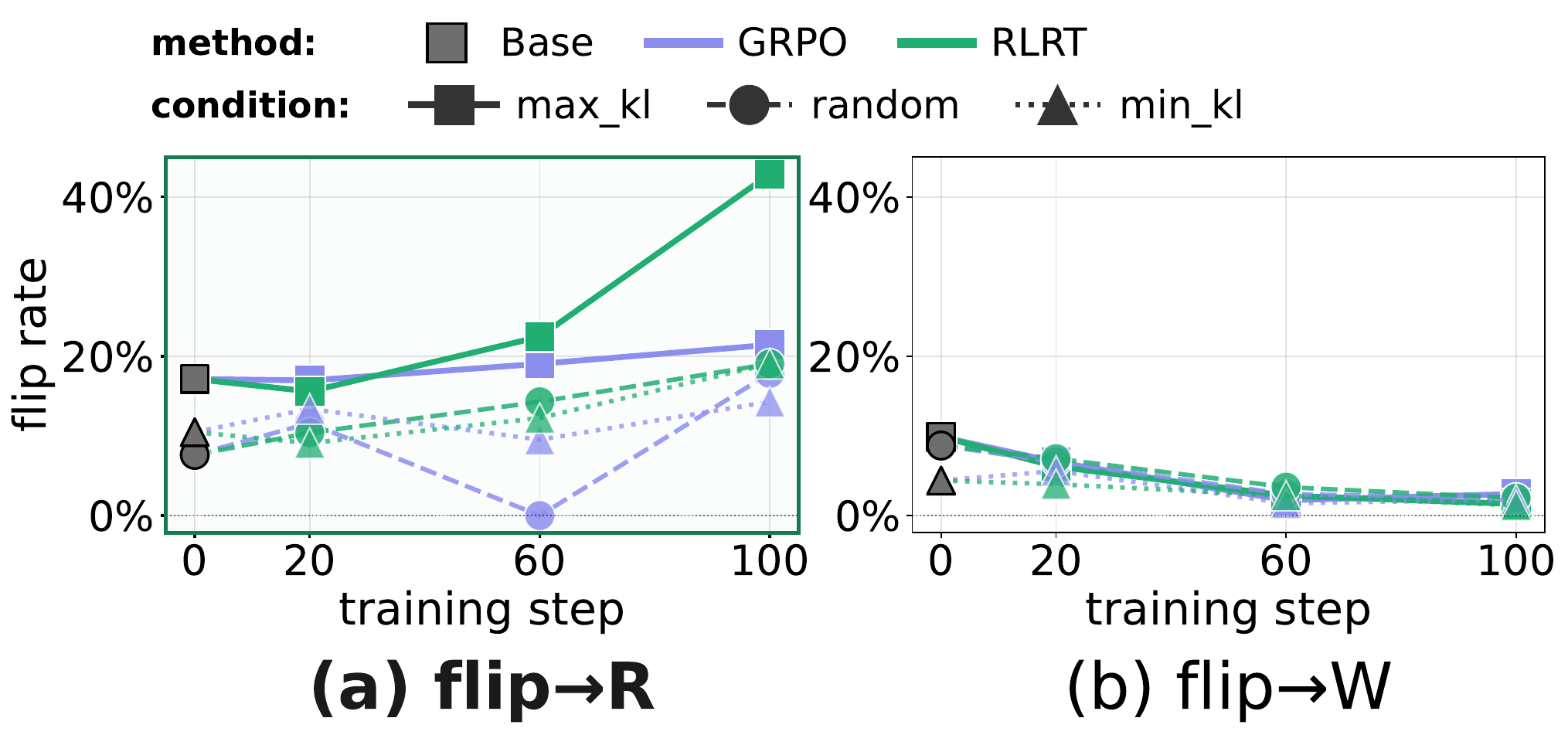}
    \caption{\small{Reflection injected at \texttt{max\_kl} 
    ($\blacksquare$), \texttt{random} ($\bullet$), or 
    \texttt{min\_kl} ($\blacktriangle$). 
    \textbf{(a)} \emph{flip$\to$R} on hard subset; 
    \textbf{(b)} \emph{flip$\to$W} on easy subset.}}
    \vspace{-0.3cm}
    \label{fig:flip-rates}
\end{wrapfigure}

Two findings emerge from Fig.~\ref{fig:flip-rates}. First, on the untuned checkpoint (step 0, $\blacksquare$), \emph{flip$\to$R} at \texttt{max\_kl} is twice that at \texttt{random} or \texttt{min\_kl}, confirming Section~\ref{sec:motiv_barDt}'s claim that $\bar{D}_t$ marks positions causally affecting correct outcomes. The absence of a comparable \emph{flip$\to$W} spike (panel b) reflects that the reflection prompt is biased toward correcting errors, though \texttt{max\_kl} remains higher than \texttt{random} and \texttt{min\_kl}. Second, the two algorithms diverge with training: \textcolor[HTML]{21AE73}{RLRT} amplifies the \texttt{max\_kl} \emph{flip$\to$R} gain from $\sim$18\% to over 40\% by step 100, while \textcolor[HTML]{8C8EEE}{GRPO} lets it collapse toward \texttt{random} and \texttt{min\_kl}. RLRT's \emph{flip$\to$W} declines just like GRPO's, so these gains do not come at the cost of fragility on correct rollouts. This explains RLRT's edge: its $\bar{D}_t$-weighted updates concentrate exploration credit on these critical positions, whereas GRPO spreads it across mostly inert tokens.

\subsection{Does RLRT Lead to More Meaningful Distributional Shifts?} \label{subsec:distribution_shift}

To answer (Q3), we analyze \emph{where} and \emph{how} each fine-tuned policy's next-token distribution $\pi_{\text{ft}}$ diverges from the base policy $\pi_{\text{base}}$, following \citet{meng2026sparse}. We focus on hard prompts ($n_\mathrm{correct}\!\in\!\{0,1,2\}$ out of $8$ under $\pi_{\text{base}}$) so that any shift reflects how the policy learns to improve on cases the base struggles with, and use $30$ such prompts from DAPO-Math-17k. At each token position along a fine-tuned rollout, we measure Jensen--Shannon divergence $\text{JS}(\pi_{\text{ft}}\,\|\,\pi_{\text{base}})$, and call positions with $\text{JS}>0.1$ \emph{high-divergence}: these are the tokens where $\pi_{\text{ft}}$ has changed its mind relative to $\pi_{\text{base}}$.

\begin{figure}[h!]
    \centering
    \vspace{-0.8em}
    \includegraphics[width=\linewidth]{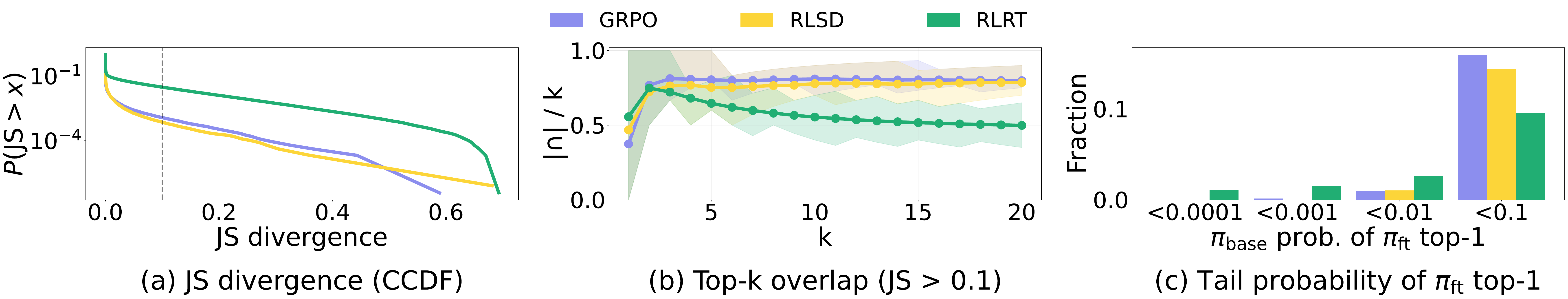}
    \caption{\small\textbf{Token-level distributional shifts of $\pi_{\text{ft}}$ relative to 
$\pi_{\text{base}}$.}
\textbf{(a)} CCDF of $\text{JS}(\pi_{\text{ft}}\|\pi_{\text{base}})$ 
across all positions; dashed line marks the $\text{JS}>0.1$ 
threshold for (b)--(c).
\textbf{(b)} Top-$k$ overlap 
$|\text{top-}k(\pi_{\text{ft}})\cap\text{top-}k(\pi_{\text{base}})|/k$ 
at high-divergence positions ($k\in[1,20]$): how much the candidate 
set is reshuffled.
\textbf{(c)} Fraction of high-divergence positions whose new top-$1$ 
token had base probability below each threshold: how deep into the tail.}
\label{fig:sparse-critical}
\end{figure}

The three panels in Figure \ref{fig:sparse-critical} answer three questions about the shift:
\begin{itemize}[itemsep=0pt, topsep=0pt, parsep=0pt, partopsep=0pt, leftmargin=*]
    \item \textbf{(a) How often does the policy diverge from the base?} Panel~(a) shows the fraction of positions with JS divergence above threshold $x$. GRPO and RLSD stay close to $\pi_{\text{base}}$ at most positions, while RLRT places far more positions in the high-divergence regime.
    \item \textbf{(b) When it diverges, do new tokens enter the top 
    candidates, or are existing ones re-ranked?} Panel~(b) measures 
    top-$k$ overlap between $\pi_{\text{ft}}$ and $\pi_{\text{base}}$ 
    at high-divergence positions. GRPO and RLSD retain $\sim80\%$ of 
    $\pi_{\text{base}}$'s candidates even at $k\geq3$, re-weighting 
    the existing pool. RLRT drops to $\sim50\%$ at $k{=}20$, 
    indicating many top candidates are tokens the base did not surface.
    \item \textbf{(c) How extreme are these new candidates?} Panel~(c) reports the fraction of high-divergence positions whose new top-$1$ token had $\pi_{\text{base}}$-probability below each threshold. RLRT promotes tokens with base probability under $10^{-3}$ to top-$1$ over $10\times$ as often as the others, routinely picking tokens the base treated as essentially zero.
\end{itemize}

Together, the three views draw a clear line.
GRPO and RLSD \emph{sharpen} what $\pi_{\text{base}}$ already prefers, re-weighting its top candidates. RLRT instead \emph{reorganizes}
the candidate set itself, pulling tokens from the base's tail into top
positions: it goes beyond reinforcing what the base knows and produces
genuinely new behavior.

\subsection{Comparison with Other Exploration Methods}

We finally answer (Q4) by comparing RLRT against two representative exploration methods: GRPO with an entropy bonus (GRPO+EB)~\citep{reasoningwithexploration} for token-level entropy regulation, and DIVER~\citep{diver} for sequence-level diversity. 

\begin{wrapfigure}{r}{0.4\textwidth}
    \centering
    \vspace{-0.4cm}
    \includegraphics[width=\linewidth]{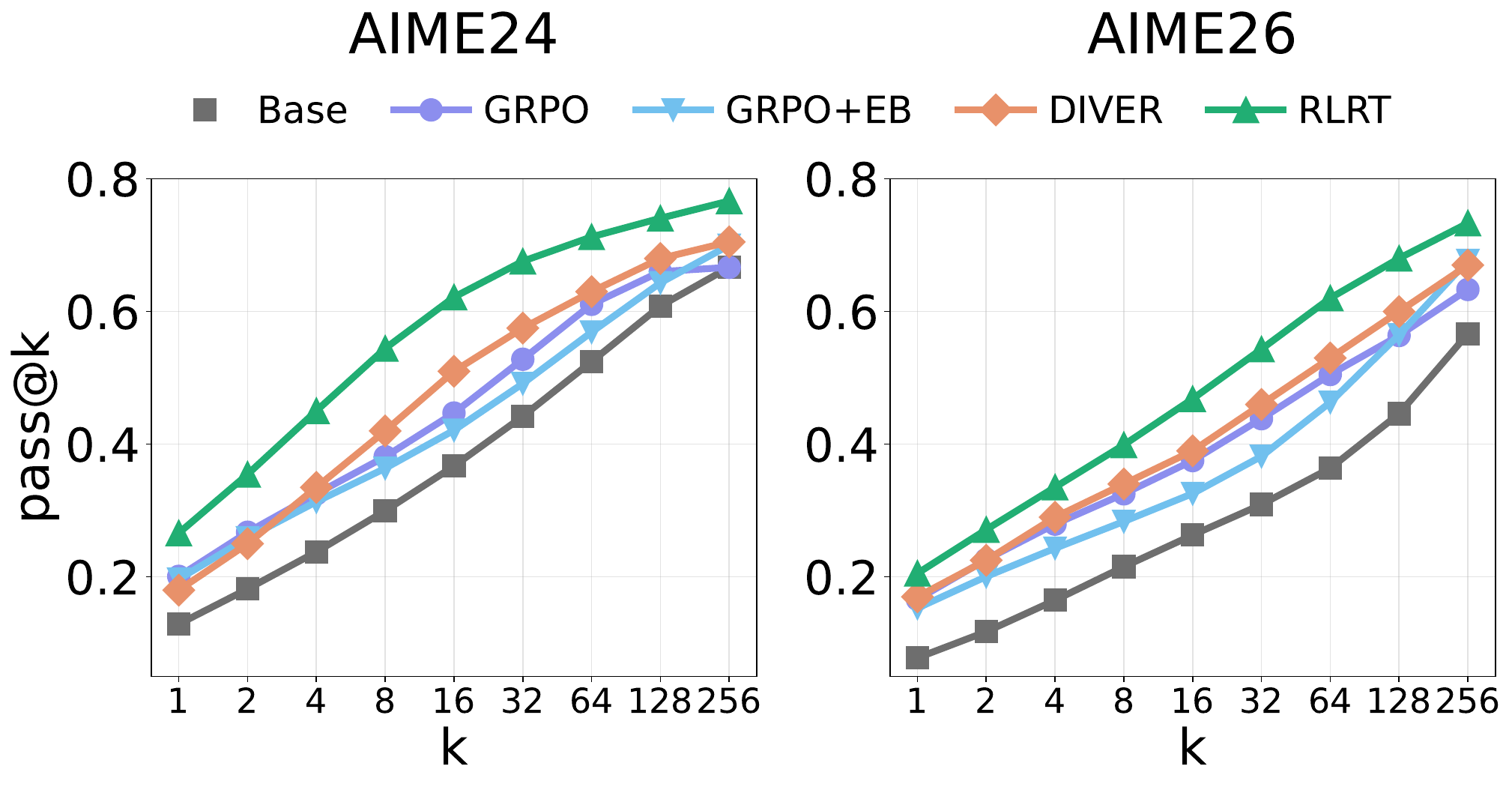}
    \vspace{-0.4cm}
    \caption{\small{Pass@$k$ comparison across exploration methods on AIME24 and AIME26.}}
    \vspace{-0.2cm}
    \label{fig:pass_at_k}
\end{wrapfigure}

For each method, we evaluate performance on Qwen3-8B-Base by comparing the pass@$k$ curve for $k \in \{1, 2, \ldots, 256\}$ on AIME24 and AIME26. We sample 256 responses per problem and compute pass@$k$ using the unbiased estimator of~\citet{chen2021evaluating}. As shown in Figure~\ref{fig:pass_at_k}, GRPO+EB injects only local stochasticity at individual decision points~\citep{diver} and tracks GRPO closely across the pass@$k$ curve, even falling below GRPO at small $k$. DIVER improves on GRPO, most visibly at large $k$, but its margin remains narrow, suggesting that its semantic-level diversity heuristic broadens exploration only modestly. RLRT, in contrast, dominates from pass@1 through pass@256, reflecting genuinely broader coverage across reasoning modes rather than within one.

\subsection{Ablation Study}

\begin{figure}[h!]
    \centering
    \includegraphics[width=\linewidth]{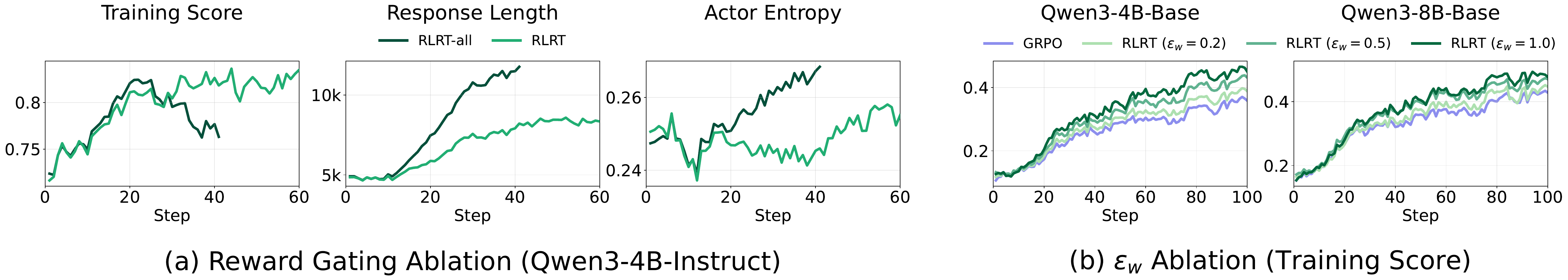}
    \caption{\small
\textbf{RLRT ablations.}
(a) reward gating on Qwen3-4B-Instruct: RLRT vs.\ RLRT-all (no $r{=}1$ gating) on training score, response length, and actor entropy.
(b) clipping range $\varepsilon_w$ on Qwen3-4B-Base and Qwen3-8B-Base, with GRPO as reference.}
\label{fig:rlrt-full-and-eps-ablation}
    \label{fig:rlrt_ablations}
\end{figure}

\paragraph{RLRT without Reward Gating.}
As described in Section~\ref{sec:method}, RLRT applies the reverse weight only on correct rollouts ($r{=}1$). To isolate the effect of the reward gate, we compare against \textbf{RLRT-all}, which applies the same weight regardless of correctness. As shown in Figure~\ref{fig:rlrt-full-and-eps-ablation} (a), RLRT-all initially tracks RLRT but then diverges: response length and entropy grow unbounded, and training collapses around step~$40$. This confirms that RLRT's gain requires restricting the reverse weight to correct rollouts: without the gate, the reverse weight reinforces teacher-divergent tokens on \emph{failed} rollouts, conflating valuable exploration with spurious divergence.

\paragraph{Effect of the Clipping Range $\varepsilon_w$.}
The clipping range $\varepsilon_w$ controls how strongly the reverse weight can deviate from $1$, and thus how much it reshapes the gradient on correct rollouts. We sweep $\varepsilon_w \in \{0.2, 0.5, 1.0\}$ on Qwen3-4B/8B-Base against a GRPO baseline. Figure~\ref{fig:rlrt-full-and-eps-ablation}~(b) shows that tighter clipping keeps the reverse weight near unity and tracks GRPO closely, while looser clipping ($\varepsilon_w = 1.0$) yields the strongest training score on both backbones. This confirms that RLRT's gains come from the reweighting itself, not from $r{=}1$ filtering alone.

\section{Conclusion}
We presented \textbf{RLRT}, which inverts self-distillation on 
correct rollouts: rather than pulling the student toward a 
privileged-context teacher, it amplifies tokens where the student 
diverged from the teacher yet still succeeded. We formalized self-driven reasoning through information asymmetry and demonstrated its effectiveness as an exploration signal both theoretically and empirically. 
Experiments on base, instruction-tuned, and thinking-tuned Qwen3 
yield substantial gains over GRPO, self-distillation, and 
exploration baselines. Extending RLRT to noisier rewards, other forms of asymmetry, and 
broader on-policy distillation beyond self-distillation, where the 
teacher distribution may come from diverse sources, is left for 
future work.

\section*{Acknowledgments}
This work was supported by Microsoft Research and in part by grants from the
Institute of Information \& Communications Technology Planning \& Evaluation
(IITP), funded by the Korea government (MSIT), under Grant No.~RS-2024-00457882
(AI Research Hub Project) and Grant No.~RS-2022-II220469 (Development of Core
Technologies for Task-oriented Reinforcement Learning for Commercialization of
Autonomous Drones).

\bibliographystyle{plainnat}
\bibliography{neurips_2026}


\newpage
\appendix

\section{Limitations and Future Directions}
\label{sec:limitations}

To our knowledge, RLRT is the first to show that reversing the teacher's signal, rather than aligning to it, can improve RLVR combined with distillation. We propose reading the information-asymmetric signal between teacher and student as a driver of exploration rather than imitation, and provide empirical evidence that this reinterpretation yields consistent gains across diverse model families. However, our setup is limited in two ways: it relies on a self-distillation framework where the teacher and student share parameters, and the experiments are restricted to mathematical reasoning.

RLRT opens several directions for future work. One axis is varying the teacher itself: rather than self-distillation, the teacher could be a separate, stronger reasoning model (as in on-policy distillation), or, conversely, a weaker one. A second axis is varying the form of privileged information given to the teacher, e.g., process-level feedback, partial hints, or failed attempts rather than a complete successful rollout. A third axis is characterizing how RLRT behaves under off-policy distillation, in contrast to the on-policy setting we study. A particularly promising direction across these axes is a hybrid that adaptively routes between teacher-guided and self-driven updates depending on the context.

\vspace{2em}

\section{RLRT Algorithm}
\label{app:rlrt-algorithm}
Algorithm~\ref{alg:rlrt} summarizes the full RLRT update. The only structural changes relative to GRPO are (i) the per-token reverse weight in Eq.~\eqref{eq:rlrt-weight} and (ii) the reward gate in Eq.~\eqref{eq:rlrt-advantage}; the rollout, reward, and trust-region mechanisms are otherwise unchanged.

\begin{algorithm}[h]
\caption{RLRT: RLVR with Reversed Teacher}
\label{alg:rlrt}
\begin{algorithmic}[1]
\REQUIRE Student/teacher $\pi_\theta$; privileged context $c$ for the teacher; prompt $x$; group size $K$; mixing $\lambda \in [0,1]$; clip radius $\varepsilon_w$.
\STATE Sample group $\{y^{(k)}\}_{k=1}^{K} \sim \pi_\theta(\cdot \mid x)$.
\STATE Compute verifiable reward $r(y^{(k)}) \in \{0,1\}$ and group-standardized advantage $A^{(k)}$ for each $k$.
\FOR{each trajectory $k = 1, \dots, K$}
    \IF{$r(y^{(k)}) = 1$}
        \FOR{$t = 1, \dots, |y^{(k)}|$}
            \STATE Compute $\hat{D}_t = \log P_S^t(y_t^{(k)}) - \log P_T^t(y_t^{(k)})$ \hfill \COMMENT{token-level information asymmetry}
            \STATE $w_t^{\mathrm{RLRT}} \leftarrow \exp\!\bigl(\mathrm{sign}(A^{(k)}) \cdot \hat{D}_t\bigr)$ \hfill \COMMENT{Eq.~\eqref{eq:rlrt-weight}}
            \STATE $A_t^{\mathrm{RLRT},(k)} \leftarrow A^{(k)} \cdot \Bigl[(1-\lambda) + \lambda \cdot \mathrm{clip}\bigl(w_t^{\mathrm{RLRT}},\, 1-\varepsilon_w,\, 1+\varepsilon_w\bigr)\Bigr]$ \hfill \COMMENT{Eq.~\eqref{eq:rlrt-advantage}}
        \ENDFOR
    \ELSE
        \STATE $A_t^{\mathrm{RLRT},(k)} \leftarrow A^{(k)}$ for all $t$ \hfill \COMMENT{vanilla GRPO advantage}
    \ENDIF
\ENDFOR
\STATE Update $\theta$ with the standard GRPO surrogate using $\{A_t^{\mathrm{RLRT},(k)}\}$.
\end{algorithmic}
\end{algorithm}

\newpage
\section{Proofs and Supporting Results}
\label{app:theory-proofs}

\subsection{Proof of Lemma~\ref{lem:bayes-teacher}}
\label{app:proof-bayes}

By the definition of $P_T^t$ and Bayes' rule,
\[
P_T^t(v) \;=\; \pi_\theta(v \mid h_t, R = 1) \;=\; \frac{\pi_\theta(R = 1 \mid h_t, y_t = v) \cdot \pi_\theta(v \mid h_t)}{\pi_\theta(R = 1 \mid h_t)}.
\]
The numerator and denominator simplify using the definitions
\[
f(v) \;:=\; \Pr[R = 1 \mid h_t, y_t = v], \qquad \bar{f}_S^t \;:=\; \mathbb{E}_{v \sim P_S^t}[f(v)] \;=\; \Pr[R = 1 \mid h_t],
\]
together with $\pi_\theta(v \mid h_t) = P_S^t(v)$, yielding
\[
P_T^t(v) \;=\; \frac{f(v) \cdot P_S^t(v)}{\bar{f}_S^t}.
\]
Taking logarithms,
\[
\log P_T^t(v) \;=\; \log f(v) + \log P_S^t(v) - \log \bar{f}_S^t.
\]
Applying the definition $\hat{D}_t(v) := \log P_S^t(v) - \log P_T^t(v)$ gives
\[
\hat{D}_t(v) \;=\; \log \bar{f}_S^t - \log f(v).
\]
\qed

\subsection{Proof of Theorem~\ref{thm:master}}
\label{app:proof-master}

The proof has two steps: Step 1 expresses $\mathrm{Inf}_S(t)$ in closed form using Lemma~\ref{lem:bayes-teacher} and bounds it by total variation distance; Step 2 applies Pinsker's inequality.

\paragraph{Step 1: bound by total variation.}
By Lemma~\ref{lem:bayes-teacher}, $f(v) = \bar{f}_S^t \cdot P_T^t(v) / P_S^t(v)$, hence
\[
f(v) - \bar{f}_S^t
\;=\; \bar{f}_S^t\!\left(\frac{P_T^t(v)}{P_S^t(v)} - 1\right)
\;=\; \frac{\bar{f}_S^t}{P_S^t(v)}\bigl(P_T^t(v) - P_S^t(v)\bigr).
\]
Substituting into the definition of $\mathrm{Inf}_S(t)$,
\begin{align*}
\mathrm{Inf}_S(t)
&\;=\; \mathbb{E}_{v \sim P_S^t}\!\bigl[\,\bigl|f(v) - \bar{f}_S^t\bigr|\,\bigr] \\
&\;=\; \sum_{v \in \mathcal{V}} P_S^t(v) \cdot \frac{\bar{f}_S^t}{P_S^t(v)}\,\bigl|P_T^t(v) - P_S^t(v)\bigr| \\
&\;=\; \bar{f}_S^t \sum_{v \in \mathcal{V}} \bigl|P_T^t(v) - P_S^t(v)\bigr| \\
&\;=\; 2\,\bar{f}_S^t \cdot \mathrm{TV}\bigl(P_S^t,\, P_T^t\bigr) \\
&\;\le\; 2\,\mathrm{TV}\bigl(P_S^t,\, P_T^t\bigr),
\end{align*}
where the second-to-last equality uses the definition $\mathrm{TV}(P, Q) := \tfrac{1}{2}\sum_v|P(v) - Q(v)|$, and the last inequality uses $\bar{f}_S^t \in [0, 1]$.

\paragraph{Step 2: Pinsker's inequality.}
For any two probability distributions $P, Q$ on a common space, Pinsker's inequality states
\[
\mathrm{TV}(P, Q) \;\le\; \sqrt{\tfrac{1}{2}\,\mathrm{KL}(P \,\|\, Q)}.
\]
Applied to $P = P_S^t$ and $Q = P_T^t$,
\[
\mathrm{TV}\bigl(P_S^t,\, P_T^t\bigr) \;\le\; \sqrt{\tfrac{1}{2}\,\bar{D}_t}.
\]
Squaring the inequality from Step~1 and applying this bound,
\[
\mathrm{Inf}_S(t)^2 \;\le\; 4\,\mathrm{TV}\bigl(P_S^t,\, P_T^t\bigr)^2 \;\le\; 4 \cdot \tfrac{1}{2}\,\bar{D}_t \;=\; 2\,\bar{D}_t,
\] 
as claimed. \qed

\section{Marker Statistics for the Explore/Exploit Reading}
\label{appendix:marker-stats}

Starting from 8 rollouts of Qwen3-8B on each of 100 DAPO-Math-17k problems, we retain one correct and one incorrect trajectory per problem (200 trajectories total). At every position $t$ of each trajectory, we identify two tokens from the entire vocabulary $\mathcal{V}$: $\arg\max_{v \in \mathcal{V}} \hat{D}_t(v)$ (the token most favored by the student over the teacher) is added to the explore corpus, and $\arg\min_{v \in \mathcal{V}} \hat{D}_t(v)$ (the token most favored by the teacher over the student) is added to the exploit corpus. Note that these are \emph{not} the sampled tokens $y_t$; they are the vocabulary entries where the student--teacher divergence is most extreme in each direction. We score every token type $v$ that appears at least 30 times across the two corpora combined, after restricting to ASCII alphabetic tokens of length 3 to 15 characters. Let $e_v, x_v$ be the counts of token $v$ in the explore/exploit corpora (totals $E, X$). We compute polarization with the smoothed log-odds $z$-score of \citet{monroe2008fightin},
\begin{equation}
  z_v = \frac{\delta_v}{\sqrt{\widehat{\mathrm{Var}}(\delta_v)}},
  \qquad
  \delta_v = \log\frac{e_v + \alpha}{E - e_v + \alpha} - \log\frac{x_v + \alpha}{X - x_v + \alpha},
\end{equation}
with $\alpha = 0.5$, where $z_v \gg 0$ marks reliable explore tokens and $z_v \ll 0$ marks reliable exploit tokens. We keep tokens with $|z_v| \geq 3$ (251 explore-side and 171 exploit-side candidates), then remove stopwords using two lists: NLTK English stopwords (198 words) and a domain-specific list (approximately 400 words covering math vocabulary, Greek letters, LaTeX fragments, tokenizer artifacts, English numerals, and generic non-discourse fillers). This yields $\mathbf{38}$ explore-side and $\mathbf{61}$ exploit-side markers, all listed in Table~\ref{tab:markers} with their $z_v$ values, ranked by $|z_v|$ within each category.

\begin{table}[h]
\centering
\small
\caption{All explore/exploit markers retained after $|z_v| \geq 3$ filtering and automatic stoplist removal, grouped by discourse function. $z_v$ is the smoothed log-odds polarization score (positive $=$ teacher-suppressed; negative $=$ teacher-favored). The ``Other'' rows list all tokens not mapping to any predefined category (13 explore, 32 exploit).}
\begin{tabular}{l l p{0.62\linewidth}}
\toprule
\textbf{Direction} & \textbf{Category} & \textbf{Tokens ($z_v$)} \\
\midrule
\multirow{9}{*}{\textcolor{custom_green_dark}{\textit{Explore}}}
  & Reflection            & \textit{wait} (33.6), \textit{back} (4.8), \textit{hmm} (3.2) \\
  & Deliberation verb     & \textit{let} (35.1), \textit{denote} (8.8), \textit{write} (8.8), \textit{note} (8.4), \textit{think} (6.1), \textit{consider} (5.5), \textit{look} (5.5), \textit{analyze} (4.9) \\
  & Metacognitive verb    & \textit{check} (12.6), \textit{recall} (7.7), \textit{see} (6.1) \\
  & Alternative marker    & \textit{another} (8.6), \textit{different} (8.3), \textit{alternative} (4.0), \textit{alternatively} (3.5) \\
  & Logical connective    & \textit{therefore} (7.0), \textit{via} (5.3), \textit{similarly} (4.7) \\
  & Contrastive           & \textit{still} (13.1) \\
  & Epistemic hedge       & \textit{maybe} (4.0), \textit{perhaps} (3.3) \\
  & Causal pivot          & \textit{since} (3.6) \\
  & Other                 & \scriptsize{\textit{answer} (9.3), \textit{icky} (6.7), \textit{ones} (5.2), \textit{problem} (5.2), \textit{contradiction} (5.1), \textit{pick} (4.4), \textit{passes} (4.4), \textit{split} (4.0), \textit{infected} (3.9), \textit{rolling} (3.9), \textit{hat} (3.8), \textit{sided} (3.8), \textit{cover} (3.7)} \\
\midrule
\multirow{9}{*}{\textcolor{custom_pink_dark}{\textit{Exploit}}}
  & Decision verb         & \textit{determine} ($-10.3$), \textit{define} ($-9.8$), \textit{apply} ($-5.3$), \textit{ensure} ($-4.9$), \textit{means} ($-4.8$), \textit{verify} ($-4.7$), \textit{yield} ($-4.6$), \textit{evaluate} ($-4.4$), \textit{compute} ($-3.8$), \textit{conclude} ($-3.2$) \\
  & Verdict adjective     & \textit{valid} ($-23.7$), \textit{incorrect} ($-11.0$), \textit{correct} ($-10.9$), \textit{right} ($-10.6$), \textit{invalid} ($-8.1$) \\
  & Consequential conn.   & \textit{thus} ($-17.0$), \textit{however} ($-6.2$), \textit{though} ($-5.0$), \textit{although} ($-3.6$) \\
  & Finalization          & \textit{final} ($-23.8$), \textit{finally} ($-19.9$), \textit{eventually} ($-8.8$) \\
  & Outcome noun          & \textit{conclusion} ($-12.9$), \textit{result} ($-10.3$), \textit{results} ($-5.2$) \\
  & Implication           & \textit{leads} ($-5.0$), \textit{implies} ($-3.1$) \\
  & Progression           & \textit{continue} ($-16.4$) \\
  & Specification         & \textit{defined} ($-4.7$) \\
  & Other                 & \scriptsize{\textit{higher} ($-16.1$), \textit{general} ($-9.7$), \textit{left} ($-9.5$), \textit{reverse} ($-8.2$), \textit{directly} ($-7.8$), \textit{always} ($-7.0$), \textit{instead} ($-6.6$), \textit{specific} ($-5.6$), \textit{text} ($-5.5$), \textit{tool} ($-4.9$), \textit{able} ($-4.5$), \textit{teams} ($-4.3$), \textit{following} ($-4.1$), \textit{guarantee} ($-4.1$), \textit{repetition} ($-3.9$), \textit{avoid} ($-3.9$), \textit{ants} ($-3.8$), \textit{cannot} ($-3.7$), \textit{trial} ($-3.7$), \textit{least} ($-3.6$), \textit{constraint} ($-3.6$), \textit{containing} ($-3.6$), \textit{actually} ($-3.6$), \textit{fully} ($-3.5$), \textit{performance} ($-3.4$), \textit{guaranteed} ($-3.3$), \textit{evenly} ($-3.3$), \textit{meaning} ($-3.2$), \textit{specifically} ($-3.1$), \textit{analysis} ($-3.1$), \textit{appear} ($-3.1$), \textit{without} ($-3.1$)} \\
\bottomrule
\end{tabular}
\label{tab:markers}
\end{table}

Varying the threshold $|z_v| \in \{2, 3, 5\}$ does not change the qualitative picture: the same discourse categories dominate on each side, and only the depth of each category's tail changes.

\section{Further Examples of Critical Positions and Explore/Exploit Directions}
\label{app:more-examples}

\begin{figure}[h]
    \centering
    \includegraphics[width=\linewidth]{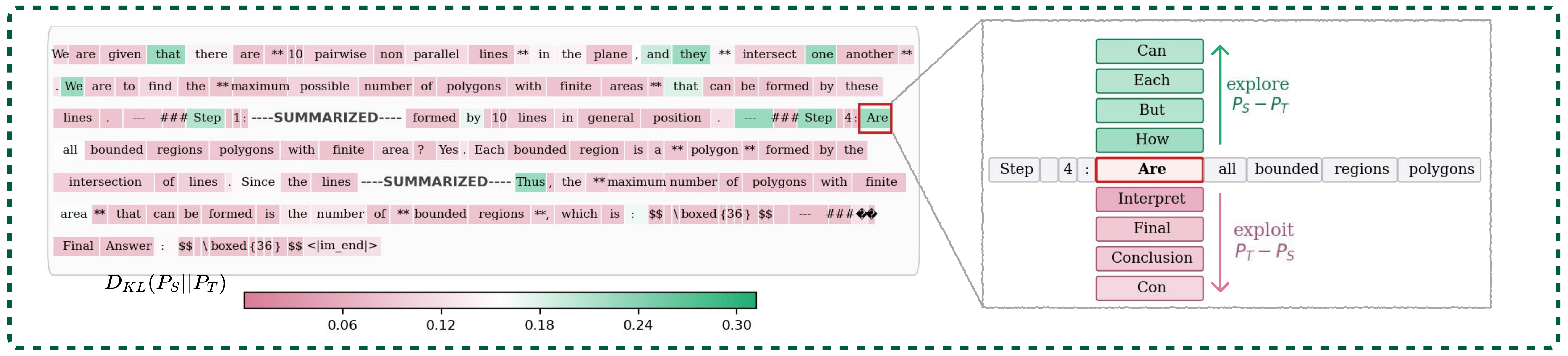}
    \caption{\textbf{Additional example of critical positions and explore/exploit directions}, complementing Figure~\ref{fig:motivation}. Token shading shows the position-level asymmetry $\bar{D}_t = \mathrm{KL}(P_S^t \,\|\, P_T^t)$. At the highlighted critical position, candidates are taken as the union of the teacher's and student's top-100 tokens; we display the top four with the largest $P_S^t - P_T^t$ (\textcolor{custom_green_dark}{green}, $\hat{D}_t > 0$, \emph{explore}) and the top four with the largest $P_T^t - P_S^t$ (\textcolor{custom_pink_dark}{pink}, $\hat{D}_t < 0$, \emph{exploit}).}
    \label{fig:more-critical-positions}
\end{figure}

Figure~\ref{fig:motivation} in Section~\ref{sec:motiv_ia} illustrates the explore/exploit decomposition on a single trajectory. To show that this pattern is not an artifact of one example, Figure~\ref{fig:more-critical-positions} presents an additional rollout annotated with the same $\bar{D}_t$ heatmap and top-candidate display. The qualitative picture replicates: most tokens carry small $\bar{D}_t$, while a few high-asymmetry tokens mark critical positions. At these positions, exploit-leaning candidates ($\hat{D}_t < 0$, e.g., \textit{Final}, \textit{Conclusion}) push toward closing the argument, whereas explore-leaning candidates ($\hat{D}_t > 0$, e.g., \textit{Can}, \textit{Each}, \textit{But}, \textit{How}) open alternative reasoning paths the teacher would not have predicted. This consistency supports the use of $\mathrm{sign}(\hat{D}_t)$ as a stable indicator of self-driven versus teacher-aligned tokens, as claimed in Section~\ref{sec:motiv_signDt}.

\newpage
\section{More Results}

\subsection{Benchmark Results on Qwen3-4B-Instruct}\label{appendix:qwen3_4b_instruct}

Extending Table~\ref{tab:main_results}, Table~\ref{tab:appendix_qwen3_4b_instruct} reports results on Qwen3-4B-Instruct across six math benchmarks (AIME24/25/26, HMMT26, AMC23, and MATH500). Consistent with the discussion in Section~\ref{sec:benchmark_results}, RLRT also achieves higher scores than other baselines on the Instruct-tuned model, yielding a 3.4\% average improvement on avg@16 over the best baseline.

\begin{table}[h!]
    \centering
    \small
    \setlength{\tabcolsep}{3pt}
    \renewcommand{\arraystretch}{1.15}
    \caption{Performance comparison on Qwen3-4B-Instruct across mathematical reasoning benchmarks. We report avg@16 and pass@16 for each benchmark. $\Delta$ denotes the gain of RLRT over the best of the other methods.}
    \resizebox{\textwidth}{!}{%
    \begin{tabular}{p{2.55cm} C{0.9cm}C{0.9cm} C{0.9cm}C{0.9cm} C{0.9cm}C{0.9cm} C{0.9cm}C{0.9cm} C{0.9cm}C{0.9cm} C{0.9cm}C{0.9cm}}
        \toprule
        \multirow{2}{*}{\textbf{Method}} 
            & \multicolumn{2}{c}{\textbf{AIME24}} 
            & \multicolumn{2}{c}{\textbf{AIME25}} 
            & \multicolumn{2}{c}{\textbf{AIME26}} 
            & \multicolumn{2}{c}{\textbf{HMMT26}} 
            & \multicolumn{2}{c}{\textbf{AMC23}} 
            & \multicolumn{2}{c}{\textbf{MATH500}} \\
        \cmidrule(lr){2-3} \cmidrule(lr){4-5} \cmidrule(lr){6-7} \cmidrule(lr){8-9} \cmidrule(lr){10-11} \cmidrule(lr){12-13}
             & {\scriptsize Avg@16} & {\scriptsize Pass@16} 
            & {\scriptsize Avg@16} & {\scriptsize Pass@16} 
            & {\scriptsize Avg@16} & {\scriptsize Pass@16} 
            & {\scriptsize Avg@16} & {\scriptsize Pass@16} 
            & {\scriptsize Avg@16} & {\scriptsize Pass@16} 
            & {\scriptsize Avg@16} & {\scriptsize Pass@16} \\
        \midrule
        \textit{Qwen3-4B-Instruct}         & 64.4 & 86.7 & 47.3 & 76.7 & 54.0 & \textbf{83.3} & 36.4 & 57.6 & 94.7 & \textbf{100.0} & 94.1 & \textbf{97.8} \\
        \quad GRPO                         & 69.4 & \textbf{93.3} & 58.3 & 80.0 & 62.7 & \textbf{83.3} & 37.9 & 60.6 & 96.2 & \textbf{100.0} & 94.4 & \textbf{97.8} \\
        \quad SDPO                         & 53.1 & 83.3 & 37.1 & 70.0 & 42.9 & 73.3 & 31.6 & 45.5 & 89.1 & \textbf{100.0} & 91.8 & 97.4 \\
        \quad RLSD                         & 61.5 & 90.0 & 51.9 & 76.7 & 58.1 & \textbf{83.3} & 37.1 & 51.5 & 93.1 & 97.5 & 94.3 & 97.2 \\
        \quad \textbf{RLRT (Ours)}         & \textbf{70.4} & 90.0 & \textbf{62.9} & \textbf{83.3} & \textbf{67.9} & \textbf{83.3} & \textbf{40.2} & \textbf{66.7} & \textbf{97.0} & \textbf{100.0} & \textbf{94.8} & \textbf{97.8} \\
        \quad $\Delta$ vs. best            & \cellcolor{gray!10}\textcolor{teal}{+1.0} & \cellcolor{gray!10}\textcolor{red!70!black}{-3.3} & \cellcolor{gray!10}\textcolor{teal}{+4.6} & \cellcolor{gray!10}\textcolor{teal}{+3.3} & \cellcolor{gray!10}\textcolor{teal}{+5.2} & \cellcolor{gray!10}{0.0} & \cellcolor{gray!10}\textcolor{teal}{+2.3} & \cellcolor{gray!10}\textcolor{teal}{+6.1} & \cellcolor{gray!10}\textcolor{teal}{+0.8} & \cellcolor{gray!10}{0.0} & \cellcolor{gray!10}\textcolor{teal}{+0.4} & \cellcolor{gray!10}{0.0} \\
        \bottomrule
    \end{tabular}%
    }
    \label{tab:appendix_qwen3_4b_instruct}
\end{table}

\vspace{2em}
\subsection{Behavior of SDPO on Base Models} \label{appendix:sdpo_in_base}

\begin{wrapfigure}{r}{0.41\textwidth}
    \centering
    \vspace{-0.35cm}
    \includegraphics[width=\linewidth]{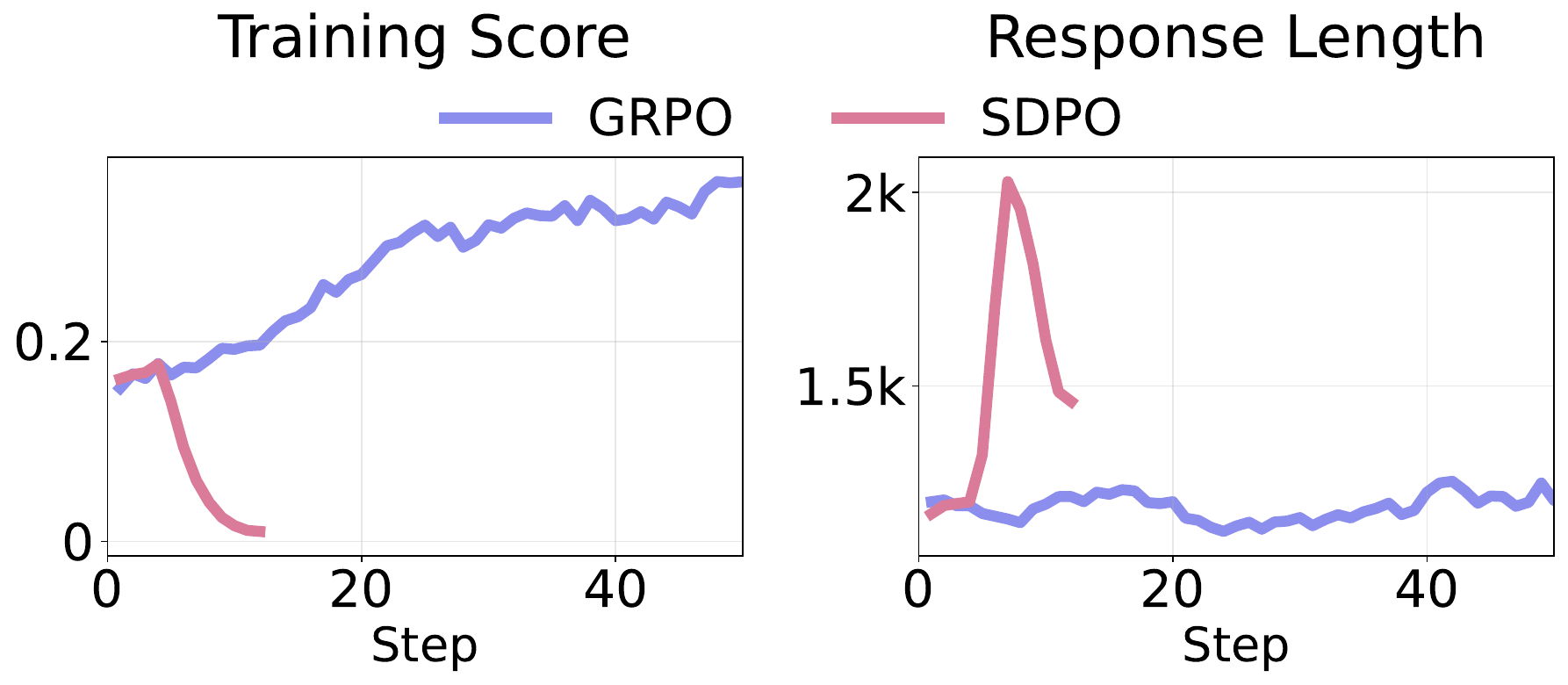}
    \caption{Training reward (left) and response length (right) on Qwen3-8B-Base. \textcolor[HTML]{DA7B99}{SDPO} collapses quickly: its reward drops while response length blows up.}
    \vspace{-0.3cm}
    \label{fig:grpo-sdpo-collapse}
\end{wrapfigure}

SDPO is a self-distillation method that uses the same model as both teacher and student under different conditioning contexts, and rapidly improves in-domain performance and induces more efficient reasoning by shortening response length \citep{sdpo}. However, it can become unstable in math reasoning due to its excessive suppression of hedging and reflective tokens (e.g., ``wait'', ``hmm''); these tokens are critical for robust reasoning \citep{why_selfdistillation}. We observe that this collapse is particularly severe on the Base model, where the score rapidly drops to 0 within 20 steps and the response length diverges compared to GRPO. We therefore exclude SDPO and its variant SRPO from the Base model comparison in our main table.

\newpage
\section{Experimental Details}
\label{app:exp_detail}
We build on the implementation of \citet{why_selfdistillation} (\url{https://github.com/beanie00/self-distillation-analysis}) and additionally implement GRPO with entropy bonus, SRPO, RLSD, and RLRT for our experiments. For DIVER \citep{diver}, we use the official code (\url{https://github.com/NJU-RL/DIVER}) and train on the same DAPO-Math-17k corpus and hyperparameters as the other baselines. We run all experiments on 2$\times$B200 GPUs. Training Qwen3-4B/8B-Base takes approximately one day, whereas Qwen3-4B-Instruct and Qwen3-8B require 2--3 days.

\subsection{Details of Baseline Algorithms} \label{appendix:baseline_details}
All baselines share the GRPO surrogate and differ only in (i) the privileged context $c$ defining the teacher view $P_T^t(\cdot) := \pi_\theta(\cdot \mid h_t, c)$, (ii) the per-token weight $w_t$ on the advantage, and (iii) the trajectory-level gate. We write $\Delta_t := \mathrm{sg}(\log P_T^t(y_t) - \log P_S^t(y_t))$ and $\hat{D}_t = -\Delta_t$ (Sec.~\ref{sec:prelim}).
\begin{itemize}[leftmargin=*,itemsep=2pt,topsep=2pt]
    \item \textbf{GRPO}~\citep{shao2024deepseekmath,dapo}. The DAPO recipe: clip-higher ($\varepsilon_{\text{low}}{=}0.2$, $\varepsilon_{\text{high}}{=}0.28$), token-level loss aggregation, no KL penalty. No teacher view.
    \item \textbf{SDPO}~\citep{sdpo}. Teacher conditions on a correct rollout; a logit-level KL loss pulls $P_S^t \to P_T^t$ on all rollouts.
    \item \textbf{SRPO}~\citep{srpo}. Same teacher as SDPO, but \emph{routed by correctness}: SDPO loss on $r{=}0$ rollouts, GRPO on $r{=}1$, with entropy-aware dynamic weighting.
    \item \textbf{RLSD}~\citep{rlsd}. Teacher conditions on the ground-truth answer. The reward fixes the update \emph{direction}, while the teacher modulates only \emph{magnitude}: $w_t^{\mathrm{RLSD}} = (P_T^t(y_t)/P_S^t(y_t))^{\mathrm{sign}(A)}$, applied to all rollouts.
\end{itemize}
We run each baseline with the primary settings recommended in its original paper.

\textbf{Relation to RLSD.} RLRT and RLSD use weights of the same form with \emph{opposite exponents}, $w_t^{\mathrm{RLRT}} = 1/w_t^{\mathrm{RLSD}}$, and RLRT additionally gates on $r{=}1$. On correct rollouts, RLSD up-weights teacher-favored tokens ($\hat{D}_t{<}0$); RLRT up-weights student-favored ones ($\hat{D}_t{>}0$), amplifying self-driven reasoning rather than imitating the teacher.

\begin{table}[h]
\centering
\small
\setlength{\tabcolsep}{4pt}
\caption{Baselines unified under the GRPO surrogate. Each method applies $A_t^{(k)} = A^{(k)} \cdot [(1-\lambda) + \lambda \cdot \mathrm{clip}(w_t, 1-\varepsilon_w, 1+\varepsilon_w)]$ and differs only in $c$, $w_t$, and the gate.}
\label{tab:baselines}
\begin{tabular}{p{2.cm}p{2.2cm}p{4cm}p{1.7cm}p{2.55cm}}
\toprule
Method & Context $c$ & Per-token weight $w_t$ & Gate & Direction \\
\midrule
GRPO  & ---            & $1$ & ---   & --- \\
SDPO  & correct rollout  & logit-level KL: $\mathrm{KL}(P_T^t \Vert P_S^t)$ & all & $P_S \!\to\! P_T$ \\
SRPO  & correct rollout  & SDPO loss if $r{=}0$;\; $1$ if $r{=}1$ & route by $r$ & $P_S \!\to\! P_T$ on $r{=}0$ \\
RLSD  & ground truth   & $\bigl(P_T^t(y_t)/P_S^t(y_t)\bigr)^{\mathrm{sign}(A)}$ & all & teacher = magnitude \\
\textbf{RLRT (ours)} & correct rollout & $\bigl(P_S^t(y_t)/P_T^t(y_t)\bigr)^{\mathrm{sign}(A)}$ & $r{=}1$ only & amplify self-driven \\
\bottomrule
\end{tabular}
\end{table}

\subsection{Hyperparameters} \label{app:hyperparams}

\paragraph{Training Hyperparameters.} The training hyperparameters are listed in Table~\ref{tab:hyperparameters_all}. For SDPO and SRPO, we follow the hyperparameter settings recommended in their original papers, sweeping only SRPO's entropy-aware dynamic-weight coefficient $\beta \in \{0, 0.5, 1\}$ per model. For RLSD and RLRT, we share $\lambda_\text{init} = 0.5$ and sweep $\epsilon_w \in \{0.2, 0.5, 1.0\}$ under an identical protocol. RLSD was consistently best with $\epsilon_w = 0.2$, with larger values degrading performance below GRPO. RLRT, by contrast, remained above GRPO across the entire sweep. The best setting shifted modestly with the base model's ability to explore diverse solution paths (base: $1.0$, instruction-tuned: $0.5$, thinking-tuned: $0.2$), consistent with the role of $\epsilon_w$ in the method.

For GRPO, we follow \citet{dr_grpo} and disable std normalization of the advantage to preserve relative signal strength across groups. For RLSD and RLRT, we retain it following RLSD \citep{rlsd}.

\begingroup
\small
\begin{longtable}{p{4cm}p{4.5cm}C{4.2cm}}
\caption{Hyperparameters for GRPO, SDPO, SRPO, RLSD, and RLRT.}
\label{tab:hyperparameters_all} \\
\toprule
\textbf{Category} & \textbf{Parameter} & \textbf{Value} \\
\midrule
\endfirsthead
\multicolumn{3}{l}{\textit{(continued from previous page)}} \\
\toprule
\textbf{Category} & \textbf{Parameter} & \textbf{Value} \\
\midrule
\endhead
\midrule
\multicolumn{3}{r}{\textit{(continued on next page)}} \\
\endfoot
\bottomrule
\endlastfoot
\multicolumn{3}{l}{\textit{Common (shared by all methods)}} \\
\midrule
\multirow{2}{*}{Data}
& Max.\ prompt length & 2048 \\
& Max.\ response length & 20480 \\
\midrule
\multirow{3}{*}{Batching}
& Question batch size & 256 \\
& Mini batch size & 128 \\
& Number of rollouts & 8 \\
\midrule
\multirow{2}{*}{Rollout}
& Inference engine & vllm \\
& Temperature & 1.0 \\
\midrule
\multirow{4}{*}{Training}
& Optimizer & AdamW \\
& Warmup steps & 10 \\
& Weight decay & 0.01 \\
& Gradient clip norm & 1.0 \\
\midrule
\multirow{3}{*}{Policy loss}
& $\epsilon$-low & 0.2 \\
& $\epsilon$-high & 0.28 \\
& Loss aggregation & token-level \\
\midrule
\multirow{2}{*}{Advantage std normalization}
& GRPO, SRPO & disabled \\
& RLSD, RLRT & enabled (Following RLSD \citep{rlsd}) \\
\midrule
\multirow{2}{*}{Off-policy correction}
& Rollout IS clip & 2 \\
& KL coefficient ($\lambda$) & 0.0 \\
\midrule
\multirow{3}{*}{Learning rate}
& GRPO / RLSD / RLRT & $1 \times 10^{-6}$ \\
& SDPO & $1 \times 10^{-5}$ \\
& SRPO & $5 \times 10^{-6}$ \\
\midrule
\multicolumn{3}{l}{\textit{SDPO / SRPO}} \\
\midrule
\multirow{4}{*}{Distillation}
& Divergence & Jensen--Shannon ($\alpha = 0.5$) \\
& Top-$K$ distillation & 100 \\
& EMA update rate & 0.0 \\
& Entropy-aware coefficient ($\beta$, SRPO only) & swept over $\{0, 0.5, 1\}$ \\
\midrule
\multicolumn{3}{l}{\textit{RLSD / RLRT}} \\
\midrule
\multirow{6}{*}{Token reweighting}
& Initial mixing ($\lambda_\text{init}$) & 0.5 \\
& $\epsilon_w$ sweep & $\{0.2, 0.5, 1.0\}$ \\
& \hspace{1em}- Best (RLSD) & $0.2$ \\
& \hspace{1em}- Best (RLRT) & $1.0$ (base), $0.5$ (instruct), $0.2$ (thinking) \\
& Mixing decay steps (RLSD) & 50 \\
& Mixing decay steps (RLRT) & no decay (base), 30 (instruct, thinking) \\
\end{longtable}
\endgroup

\paragraph{Evaluation Hyperparameters.} Following the evaluation recommendations for each model\footnote{\url{https://huggingface.co/Qwen/Qwen3-4B-Instruct-2507}, \url{https://huggingface.co/Qwen/Qwen3-8B}}, we use a maximum response length of 38912 tokens, temperature 0.7, top-$p = 0.8$, and top-$K = 20$ across all models.

\clearpage
\section{Full-Trajectory Heatmaps of $\bar{D}_t$}
\label{app:full_heatmap}
The figure in Section~\ref{sec:motivation} highlights a single critical position per rollout. For completeness, we provide full-trajectory heatmaps of the position-level information asymmetry $\bar{D}_t = \mathrm{KL}(P_S^t \,\|\, P_T^t)$ across entire rollouts. Each token is shaded by its $\bar{D}_t$ value: \textcolor{custom_green_dark}{greener} tokens are critical (token choice can change correctness), while \textcolor{custom_pink_dark}{pinker} stretches are routine. The heatmaps reveal two qualitative properties of the signal that the zoomed-in view cannot convey: (i)~critical positions are sparse and concentrated, with the bulk of any rollout consisting of decision-insensitive tokens, and (ii)~they cluster at semantically meaningful junctions such as step transitions, choice of solution strategy, and arithmetic commitments, rather than scatter uniformly.

\begin{figure}[h!]
    \centering
    \includegraphics[width=\linewidth]{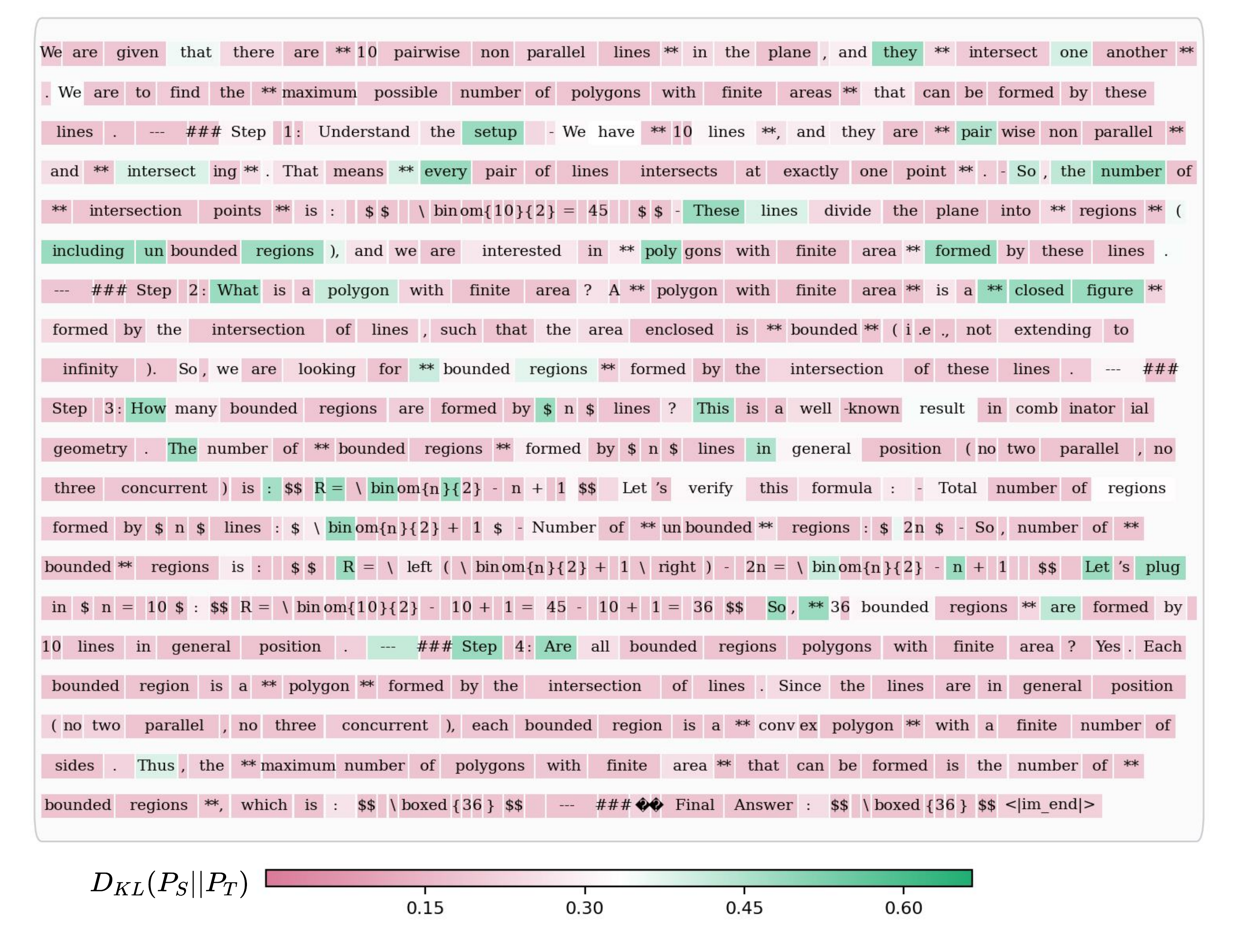}
    \caption{Full-trajectory heatmap of $\bar{D}_t$ on the first example rollout. Critical positions (\textcolor{custom_green_dark}{green}) are sparse and concentrate at decision points, while long routine stretches (\textcolor{custom_pink_dark}{pink}) carry little signal.}
    \label{app:prob1_full_heatmap}
\end{figure}

\begin{figure}[h!]
    \centering
    \includegraphics[width=\linewidth]{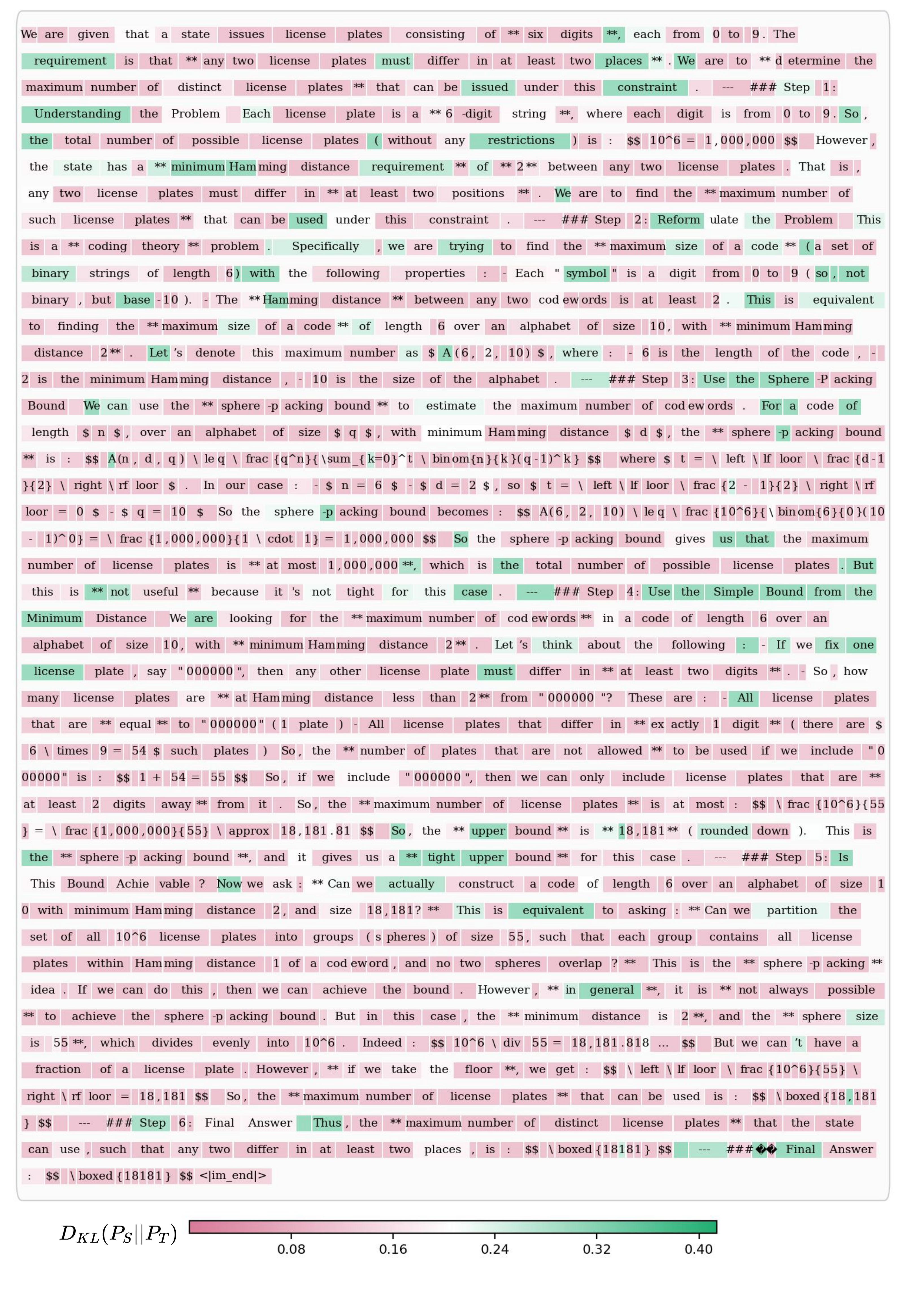}
    \caption{Full-trajectory heatmap of $\bar{D}_t$ on the second example rollout (same conventions as Figure~\ref{app:prob1_full_heatmap}).}
    \label{app:prob2_full_heatmap}
\end{figure}


\end{document}